\RequirePackage{fix-cm}
\RequirePackage{amsmath}

\documentclass{article}
\usepackage{natbib}
\usepackage[left=3.5cm, right=3.5cm, top=3.5cm, bottom=3.5cm]{geometry}
\usepackage{authblk}

\usepackage{amssymb, amsfonts}
\usepackage{graphicx}
\usepackage[algoruled]{algorithm2e}
\usepackage{bbm}
\usepackage[scaled=0.85]{beramono}    
\usepackage[utf8]{inputenc} 
\usepackage[T1]{fontenc}    
\usepackage{hyperref}       
\hypersetup{%
     colorlinks   = true,
     citecolor    = blue
}
\usepackage{url}            

\usepackage{subcaption}
\usepackage[labelfont=bf,labelsep=space]{caption}
\usepackage{tikz}
\usepackage{float}
\usepackage{enumerate}

\newcommand{\myunit}{2pt} 
\newcommand\new[1]{\textcolor{black}{#1}}

\newcommand{\bs}[1]{\boldsymbol{\mathbf{#1}}}

\DeclareMathOperator*{\argmin}{arg\,min}
\DeclareMathOperator*{\argmax}{arg\,max}

\usetikzlibrary{fit,positioning,matrix}
\usetikzlibrary{matrix,arrows,decorations.pathmorphing}
\tikzset{
    node style  one/.style={draw,fill=black!60, circle,minimum size=\myunit},
    node style  zero/.style={draw,circle,minimum size=\myunit},
    node style  blue/.style={draw,fill=blue!60, circle,minimum size=\myunit},
    node style  red/.style={draw,fill=red!60, circle,minimum size=\myunit},
    node style  green/.style={draw,fill=green!60, circle,minimum size=\myunit},
    node style  na/.style={ draw,fill=white, circle,dashed, inner sep=1.3mm,minimum size=\myunit},
    node style    sp/.style={draw,circle,minimum size=\myunit},
    node style    ge/.style={circle,minimum size=\myunit},
    arrow style  mul/.style={draw,sloped,midway,fill=white},
    arrow style plus/.style={midway,sloped,fill=white}
}

\usepackage{ulem}
\newcommand{\ced}[1]{\textcolor{black}{#1}}
\newcommand{\ve}[1]{ {\mathbf{#1}} }

\begin{document}
\sloppy

\title{Bayesian Mean-parameterized Nonnegative Binary Matrix Factorization}


\author[1]{Alberto Lumbreras}
\author[2]{Louis Filstroff}
\author[2]{Cédric Févotte}
\affil[1]{Criteo AI Lab, France}
\affil[2]{IRIT, Université de Toulouse, CNRS, France}

\maketitle

\begin{abstract}
Binary data matrices can represent many types of data such as social networks, votes, or gene expression. In some cases, the analysis of binary matrices can be tackled with nonnegative matrix factorization (NMF), where the observed data matrix is approximated by the product of two smaller nonnegative matrices. In this context, probabilistic NMF assumes a generative model where the data is usually Bernoulli-distributed. Often, a link function is used to map the factorization to the $[0,1]$ range, ensuring a valid Bernoulli mean parameter. However, link functions have the potential disadvantage to lead to uninterpretable models.
Mean-parameterized NMF, on the contrary, overcomes this problem. We propose a unified framework for Bayesian mean-parameterized nonnegative binary matrix factorization models (NBMF). We analyze three models which correspond to three possible constraints that respect the mean-parameterization without the need for link functions. Furthermore, we derive a novel collapsed Gibbs sampler and a collapsed variational algorithm to infer the posterior distribution of the factors. Next, we extend the proposed models to a nonparametric setting where the number of used latent dimensions is automatically driven by the observed data. We analyze the performance of our NBMF methods in multiple datasets for different tasks such as dictionary learning and prediction of missing data. Experiments show that our methods provide similar or superior results than the state of the art, while automatically detecting the number of relevant components.
\end{abstract}

\section{Introduction}

Nonnegative matrix factorization (NMF) is a family of methods that approximate a nonnegative matrix $\mathbf{V}$ of size ${F \times N}$ as the product of two nonnegative matrices,
\begin{align}
\bs{V} \approx \bs{WH},
\end{align}
where $\mathbf{W}$ has size $F \times K$, and $\mathbf{H}$ has size ${K \times N}$, often referred to as the {\em dictionary} and the {\em activation matrix}, respectively. $K$ is usually chosen such that $FK + KN \ll~FN$, hence reducing the data dimension.

Such an approximation is often sought after by minimizing a measure of fit between the observed data $\mathbf{V}$ and its factorized approximation $\mathbf{WH}$, i.e., 
\begin{align}
\bs{W}, \bs{H} = \argmin_{\bs{W,H}} D(\bs{V} | \bs{WH}) \quad \text{s.t} \quad \bs{W} \geq 0, \quad \bs{H} \geq 0,
\end{align}
where $D$ denotes the cost function, and where the notation $\bs{A} \geq 0$ denotes nonnegativity of the entries of $\bs{A}$.
 Typical cost functions include the squared Euclidean distance and the generalized Kullback-Leiber divergence \citep{Lee1999}, the $\alpha$-divergence \citep{Cichocki2008} or the $\beta$-divergence \citep{Fevotte2011}. Most of these cost functions underlie a probabilistic model for the data, such that minimization of the cost function is equivalent to  joint maximum likelihood estimation of the factors \citep{Singh2008}, i.e.,
\begin{align}
\argmin_{\bs{W}, \bs{H}} D(\bs{V} | \bs{WH}) = \argmax_{\bs{W,H}} p(\bs{V} | \bs{W},\bs{H}),
\end{align}
where $p$ is a probability distribution. As such, so-called \textit{Bayesian NMF} can be considered, where the factors $\mathbf{W}$ and $\mathbf{H}$ are assumed to be random variables with prior distributions, and inference is based on their posterior distribution, i.e.,
\begin{align}
p(\bs{W}, \bs{H} | \bs{V}) = p(\bs{V} | \bs{W}, \bs{H})p(\bs{W}, \bs{H})/ p(\bs{V}).
\end{align}
This has notably been addressed for different models such as Poisson \citep{Cemgil2009}, additive Gaussian \citep{Schmidt2009,Alquier2017}, or multiplicative Exponential \citep{Hoffman2010a}.

In this paper, we are interested in Bayesian NMF for binary data matrices. Binary matrices may represent a large variety of data such as social networks, voting data,  gene expression data, or binary images. 
As we shall see in Section~\ref{sec:soa}, a common practice is to consider the model
\begin{equation}
p(\mathbf{V}|\mathbf{W},\mathbf{H}) = \prod_{f,n} \text{Bernoulli} \left( v_{fn} | \phi ([\mathbf{WH}]_{fn}) \right),
\label{eq:bern-link}
\end{equation}
where $\phi$ is a link function that maps the factorization $\mathbf{WH}$ to the $[0,1]$ range.\footnote{Distributions used throughout the article are formally defined in Appendix~\ref{app:dist}.} 
Although link functions are convenient since they allow the factors to be unconstrained, and sometimes result in tractable problems, they sacrifice the mean-parameterization of the Bernoulli likelihood (i.e. $\mathbb{E}[\mathbf{V}|\mathbf{WH}] = \phi(\mathbf{WH})$ instead of $\mathbb{E}[\mathbf{V}|\mathbf{WH}] = \mathbf{WH}$).

Mean-parameterized nonnegative binary matrix factorization (NBMF), however, does not rely on a link function ---or equivalently, considers $\phi(\bs{WH}) = \bs{WH}$--- and assumes the likelihood of the data to be
\begin{equation*}
p(\mathbf{V}|\mathbf{W},\mathbf{H}) = \prod_{f,n} \text{Bernoulli}(v_{fn}|[\mathbf{WH}]_{fn}),
\end{equation*}
which implies $\mathbb{E}[\mathbf{V}|\mathbf{WH}] = \mathbf{WH}$. 
Mean-parameterization is an interesting property of a model because it makes the decomposition easy to interpret. Besides, in a Bernoulli likelihood, the product  $\mathbf{WH}$ ---and, in this paper, the individual factors as well--- can be interpreted as probabilities. An additional advantage of dealing with probabilities is that they lay in a linear, continuous space, where we can apply off-the-shelf clustering methods over the latent factors. For instance, in recommender systems, we may want to cluster users by latent musical preferences, or by the latent type of product they buy. Or we may want a user to see only those categories with a probability higher than some threshold.

Our contributions in this paper are the following:
\begin{enumerate}[(a)] 
\item we present a unified framework for three Bayesian mean-parameterized NBMF models that place three possible constraints on the factors;
\item we derive a collapsed Gibbs sampler as well as collapsed variational inference algorithms, which have never been considered for these models;
\item we discuss the extension of the models to a nonparametric setting ---where the number of latent components does not need to be fixed a priori--- and propose an approximation that shows excellent results with real data.
\end{enumerate}

We test the performance of the models for different tasks in multiple datasets and show that our models give similar or superior results to the state of the art, while automatically detecting the number of relevant components.
The datasets, the algorithms, and scripts to replicate all the reported results are available through an R package. \footnote{\url{https://github.com/alumbreras/NBMF}}

\section{Related work}\label{sec:soa}

\subsection{Logistic PCA family \label{sec:logPCA}}

One of the earliest probabilistic approaches to model binary data matrices comes from PCA-related methods. The reformulation of PCA as a probabilistic generative model with a Gaussian likelihood \citep{Sammel1997, Tipping1999} opened the door to considering other likelihoods such as Bernoulli models, which are more appropriate for binary observations. We refer to it as logistic PCA. The maximum likelihood estimator in the model is given by 
\begin{align}
v_{fn} \sim \text{Bernoulli}\left(\sigma\left([\ve{W} \ve{H}]_{fn}\right)\right),
\end{align}
where $\sigma$ is the logistic function $\sigma(x) = 1/(1+e^{-x})$. Note that in this model 
the expectation is a non-linear transformation of the factors, such that $\mathbb{E}[\bs{V} | \bs{WH}] = \sigma(\bs{WH})$.

There are multiple maximum likelihood estimation algorithms for logistic PCA. 
For instance, while \cite{Sammel1997} use a Monte-Carlo Expectation Minimization (MC-EM) algorithm,  \cite{Tipping1998} derives a faster variational EM (vEM) algorithm. \cite{Collins2001} generalize probabilistic PCA to the exponential family and propose a general algorithm that exploits the duality between likelihoods in the exponential family and Bregman divergences.  Later, \cite{Schein2003} improved the algorithm of  \cite{Collins2001}, thanks to the optimization of a tight upper bound by Alternate Least Squares (ALS). Finally, note that inference could also be tackled with Polya-Gamma data augmentation schemes \cite{Polson2012}.

Other models similar to logistic PCA have been proposed with various priors or constraints over the factors. Some examples are \cite{Lobato2014}, where the factors are given Gaussian priors, \cite{Tome2013}, which allows one factor to have negative values, and \cite{Larsen2015}, where both factors are nonnegative. \cite{meeds2007modeling} consider the same logistic link function but a three factor decomposition $\sigma(\mathbf{WXH}$), where $\mathbf{W}$ and $\mathbf{H}$ are binary factors that represent cluster assignments, and $\mathbf{X}$ is a real-valued matrix that encodes the relations between the clusters. The expectation in these models is always $\mathbb{E}[\bs{V} | \bs{WH}] = \sigma(\bs{WH})$ or $\mathbb{E}[\bs{V} | \bs{WXH}] = \sigma(\bs{WXH})$.

\subsection{Poisson matrix factorization \label{sec:PMF}}

For practical reasons, some works have considered Poisson matrix factorization (PMF) techniques for binary data. In this case the binary nature of the data is ignored and a Poisson likelihood is considered:
\begin{align}
v_{fn} \sim \text{Poisson}([\ve{W} \ve{H}]_{fn}).
\end{align}

Different flavors of PMF have been proposed, in frequentists or Bayesian settings, and can be found, for example, in \cite{Lee1999, Canny2004, Cemgil2009, Zhou2012, Gopalan2014, Gopalan2015}. 
An advantage of PMF is that it is mean-parameterized, i.e., $\mathbb{E}[\bs{V} | \bs{WH}] = \bs{WH}$. Another useful advantage is that inference algorithms need only iterate over non-zero values, which makes them very efficient for sparse matrices. In our case, a significant disadvantage is that it assigns non-zero probabilities to impossible observations ($v_{fn} > 1)$.

\begin{figure}[t]
\centering
\subcaptionbox{Identity}{
\includegraphics[width=0.3\textwidth]{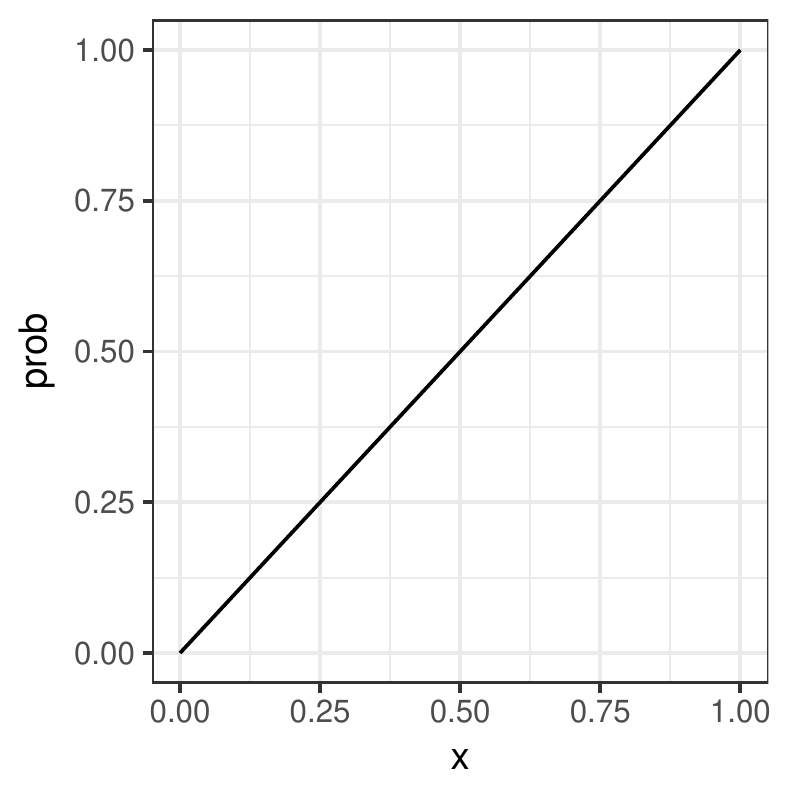}
}
\subcaptionbox{Logistic}{
\includegraphics[width=0.3\textwidth]{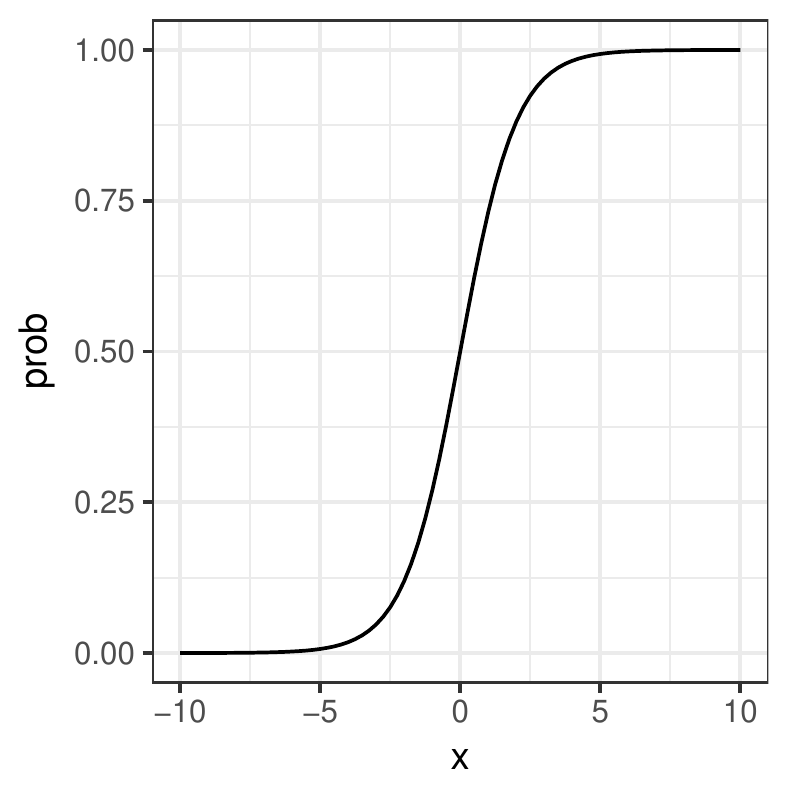}
}
\subcaptionbox{Bernoulli-Poisson}{
\includegraphics[width=0.3\textwidth]{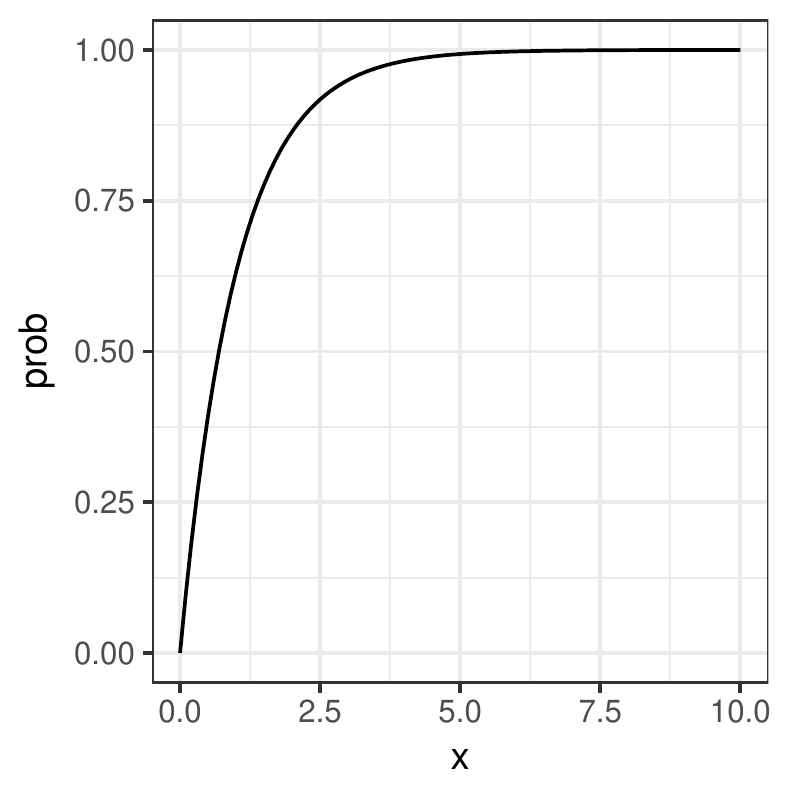}
}
\caption{Three possible link functions (identity induces mean-parameterization). }
\label{fig:linkfunctions}
\end{figure}

As we discussed above, a more reasonable choice consists in replacing the Poisson distribution with a Bernoulli distribution, possibly using some link function that maps the parameter into a $[0,1]$ range, ensuring a valid Bernoulli parameter. 
Unfortunately, unlike Poisson models, zeroes and ones under Bernoulli likelihoods do not represent counts but classes ---a zero can be considered as a \textit{no}, while a one can be considered as a \textit{yes}--- and the algorithms need to iterate over all the elements of the observation matrix. To bypass this, \cite{Zhou15, Zhou2016} proposed using the alternative link function $f(x)= 1 - e^{-x}$, coined Bernoulli-Poisson, such that
\begin{align}
v_{fn} \sim \text{Bernoulli}\left(f\left([\ve{W} \ve{H}]_{fn}\right)\right).
\end{align}
Thanks to the new link function, the model can be ``augmented'' to a Poisson model by introducing latent variables $c_{fn}$ such that
\begin{align}
c_{fn} &\sim \text{Poisson}([\ve{W} \ve{H}]_{fn})\\
v_{fn} &= \mathbbm{1}[c_{fn} \geq 1],
\end{align}
where $\mathbbm{1}$ is the set indicator function. By placing conjugate Gamma priors over the factors, posteriors can be obtained by Gibbs sampling. Expectation in this model is $\mathbb{E}[\bs{V} | \bs{WH}] = f(\bs{WH})$. Figure \ref{fig:linkfunctions} shows the logistic and Bernoulli-Poisson functions. Note that each link function has a different input domain, leading to different priors or constraints over the factors.

\subsection{Bernoulli mean-parameterized matrix factorization \label{sec:bNMF}}
None of the above methods offers mean-parameterization \textit{and} assumes a Bernoulli distribution over the data. The family of \textit{mean-parameterized Bernoulli models} is the basis of the models presented in this paper. These models have first been introduced in \cite{Kaban2008} (binary ICA) and \cite{Bingham2009} (Aspect Bernoulli model) where the constraints $\sum_k w_{fk} = 1, h_{kn} \in [0,1]$, or vice versa, are imposed on the factors.  The trick is that these constraints induce convex combinations of binary elements such that $\sum_k w_{fk} h_{kn} \in [0,1]$, which gives a valid Bernoulli parameter. 
In \cite{Bingham2009} the constraints are imposed explicitly, and the maximum likelihood estimator is computed by using EM with an augmented version of the model. In  \cite{Kaban2008} the constraint is imposed through Dirichlet and Beta priors over the factors, and Variational Bayes (VB) estimation of their posteriors is derived exploiting a similar augmentation scheme.

Table \ref{tab:soa} presents a summary of the methods presented in Sections~\ref{sec:logPCA}-\ref{sec:bNMF} that use a Bernoulli likelihood.

\begin{table}[t]
\caption{Bernoulli matrix factorization methods considered in the literature. Bernoulli, Gamma, Dirichlet distributions are denoted as Ber, Ga, and Dir, respectively. Gradient refers to Gradient-based optimization.}
\label{tab:soa}
\begin{tabular}{lllll}
\hline\noalign{\smallskip}
Reference & Likelihood  & Prior / Constr. & Estimation \\
\noalign{\smallskip}\hline\noalign{\smallskip}
\cite{Sammel1997}&
$\text{Ber}(\sigma([\bs{WH}]_{fn}))$ & 
$w_{fk} \sim \text{Normal}(\cdot)$ & 
MC-EM \\
&
& 
$h_{kn} \in \mathbb{R}$ & \\
\cite{Tipping1998} &
$\text{Ber}(\sigma([\bs{W}\bs{H}]_{fn}))$ &
$w_{fk} \sim \text{Normal}(\cdot)$ &  
vEM\\
&
& 
$h_{kn} \in \mathbb{R}$ & \\
\cite{Collins2001} &
$\text{Ber}(\sigma([\bs{W}\bs{H}]_{fn}))$ & 
$w_{fk} \in \mathbb{R}$ & Gradient\\
&
& 
$h_{kn} \in \mathbb{R}$ & \\
 \cite{Schein2003} &
$\text{Ber}(\sigma([\bs{W}\bs{H}]_{fn}))$ &
 $w_{fk} \in \mathbb{R}$ & ALS \\
&
&
 $h_{kn} \in \mathbb{R}$ & \\
 \cite{meeds2007modeling} &
$\text{Ber}(\sigma([\bs{W}\bs{X}\bs{H}]_{fn}))$ & 
$w_{fk}\sim \text{Ber}(\cdot)$ 
 & 
 Gibbs\\
&
&
$h_{kn} \sim \text{Ber}(\cdot)$ 
 & 
 \\
 \cite{Kaban2008} &
$\text{Ber}([\bs{W}\bs{H}]_{fn})$ & 
$w_{kn} \sim \text{Beta}(\cdot)
$ & VB
\\ 
&
&
$\bs{h}_{n} \sim \text{Dir}(\cdot)$ &
\\
 \cite{Bingham2009} &
$\text{Ber}([\bs{W}\bs{H}]_{fn})$ & 
$\sum_k w_{fk} =1$  & EM
\\ 
&
& 
$h_{kn} \in [0,1]$ 
&
\\ 
\cite{Tome2013} &
$\text{Ber}(\sigma([\bs{W}\bs{H}]_{fn}))$ & 
$w_{fk} \in \mathbb{R}_+$ & Gradient\\
&
& 
$h_{kn} \in \mathbb{R}$ & \\
\cite{Larsen2015}
 &
$\text{Ber}(\sigma([\bs{W}\bs{H}]_{fn}))$ & $w_{fk} \in \mathbb{R}_+$ & Gradient\\
&
& $h_{kn} \in \mathbb{R}_+$ &\\
\cite{Zhou15}&
$\text{Ber}(f([\bs{W}\bs{H}]_{fn}))$ &
$w_{fk} \sim \text{Ga}(\cdot)$
&
Gibbs
\\
&
&
$h_{kn} \sim \text{Ga}(\cdot)$
&
\\
\noalign{\smallskip}\hline
\end{tabular}
\end{table}

\subsection{Others}
Some models have also been proposed to find binary decompositions, that is, matrix factorizations where $\bs{W}$ and $\bs{H}$ contain binary elements. For instance, \cite{Zhang2010} aim to minimize a Euclidean distance or, equivalently, maximize a Gaussian likelihood under the binary constraint. Similarly, the ``discrete basis problem'' of \cite{miettinen2008discrete} aims to minimize a $L1$-norm under the same constraint. In \cite{slawski2013matrix}, an algorithm is proposed to retrieve the exact factorization when one of the factors is constrained to be binary, and the other one to be stochastic, i.e., $\sum_k h_{kn} = 1$. More recently  \cite{rukat2017} proposed a Bayesian model for the Boolean matrix factorization problem, where $\bs{W}\bs{H}$ is a Boolean product. \cite{Capan2018} proposed sum-conditioned Poisson factorization models that apply to binary data.

\ced{In the general model defined by Eq.~\eqref{eq:bern-link}, we are essentially assuming that $\ve{V} \approx \phi(\bs{W}\bs{H})$. This can be seen as a one-layer generative network with input $\bs{H}$, weight $\bs{W}$ and non-linearity $\phi(\cdot)$. As such it is possible to conceive more general models by stacking various layers or modeling $\bs{W}$ and $\bs{H}$ as the outputs of deep networks themselves. In the context of recommendation systems, where binary matrices can represent either binary ratings or implicit feedback, this has been for example considered in \cite{Xiangnan17, HongJian17}.}

\section{Mean-parameterized Bernoulli models}

\subsection{\ced{Models}}

Let us consider a mean-parameterized Bernoulli model for an observed binary matrix $\mathbf{V}$ and two latent factors $\mathbf{W}$, $\mathbf{H}$:
\begin{equation}
v_{fn} \sim \text{Bernoulli}([\bs{WH}]_{fn}). 
\label{eq:bern-nolink}
\end{equation}
To guarantee valid Bernoulli parameters, we can impose three possible sets of constraints on the factors such that $\sum_k w_{fk} h_{kn} \in [0,1]$:

\begin{center}
\begin{tabular}{ p{3cm}  p{3cm}  p{3cm} }
\text{(c1)} & \text{(c2)} & \text{(c3)} \\[0.7em]
$h_{kn} \in [0,1]$ & $\sum_k h_{kn} = 1$ & $\sum_k h_{kn} = 1$ \\[0.7em]
$\sum_k w_{fk} = 1$ & $w_{fk} \in [0,1]$ & $\sum_k w_{fk} = 1$
\end{tabular}
\end{center}

In a Bayesian setting, we may place Beta and Dirichlet priors over the factors to respect these constraints:
\begin{align*}
\begin{split}
& \texttt{Beta-Dir} \text{~(c1)} \\
& h_{kn} \sim \text{Beta}(\alpha_k, \beta_k)\\
& \mathbf{w}_{f} \sim \text{Dirichlet}(\boldsymbol{\gamma})
\end{split}
\quad
\begin{split}
& \texttt{Dir-Beta} \text{~(c2)} \\
& \bs{h}_{n} \sim \text{Dirichlet}(\bs{\eta})\\
& w_{fk} \sim \text{Beta}(\alpha_k, \beta_k)
\end{split}
\qquad
\begin{split}
& \texttt{Dir-Dir} \text{~(c3)} \\
& \mathbf{h}_{n} \sim \text{Dirichlet}(\boldsymbol{\eta}) \\
& \mathbf{w}_{f} \sim \text{Dirichlet}(\boldsymbol{\gamma})
\end{split}
\end{align*}
where $\mathbf{h}_n$ denotes the $n$-th column of the matrix $\mathbf{H}$, and $\mathbf{w}_f$ denotes the $f$-th row of the matrix $\mathbf{W}$. The Beta parameters are positive real numbers $\alpha_k, \beta_k \in \mathbb{R}_{++}$ and the Dirichlet parameters are $K$-dimensional vectors of positive real numbers $\bs{\gamma}, \bs{\eta}  \in \mathbb{R}_{++}^{K}$.

\begin{figure}[t]
\centering
\subcaptionbox{\texttt{Beta-Dir}}{
\vspace{-0.5cm} 
\includegraphics[width=0.25\textwidth]{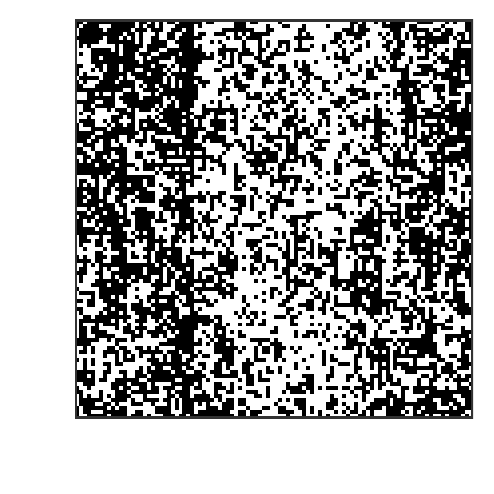}
\includegraphics[width=0.25\textwidth]{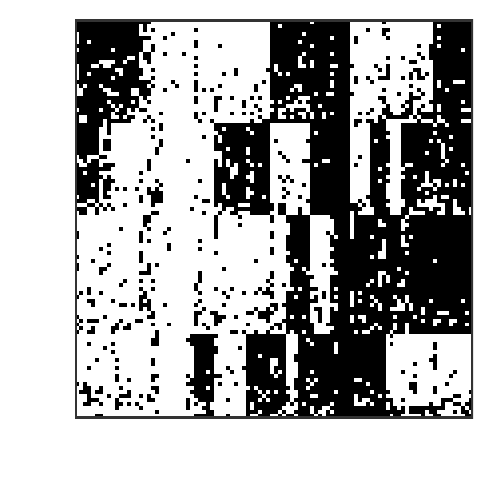}
}
\subcaptionbox{\texttt{Dir-Beta}}{
\vspace{-0.5cm} 
\includegraphics[width=0.25\textwidth]{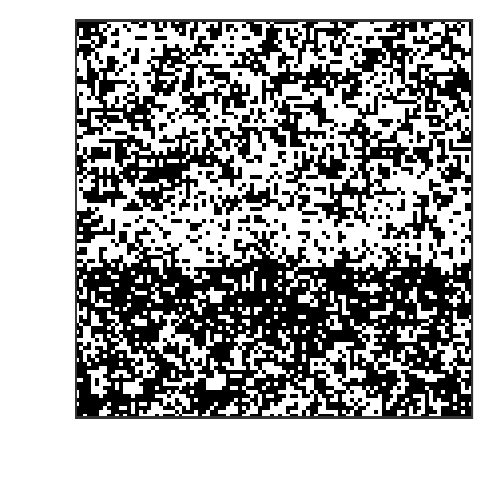}
\includegraphics[width=0.25\textwidth]{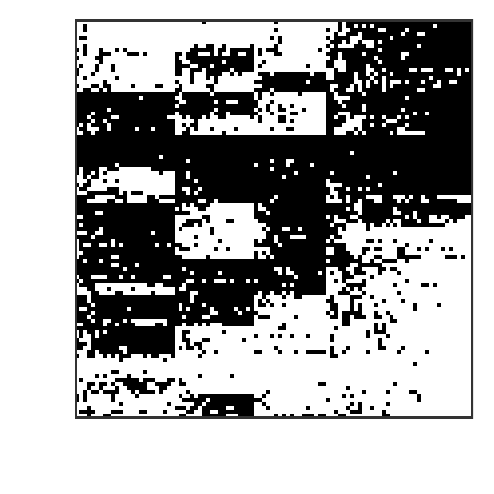}
}
\subcaptionbox{\texttt{Dir-Dir}}{
\vspace{-0.5cm} 
\includegraphics[width=0.25\textwidth]{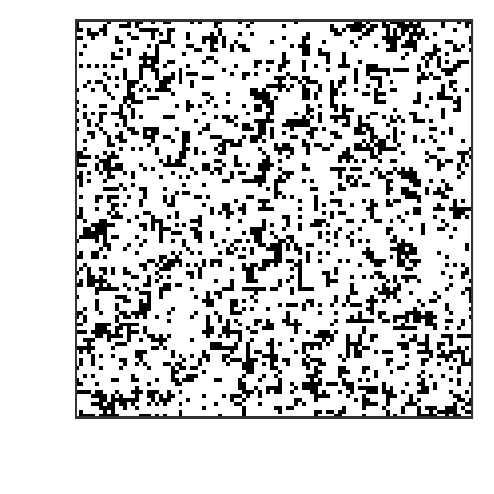}
\includegraphics[width=0.25\textwidth]{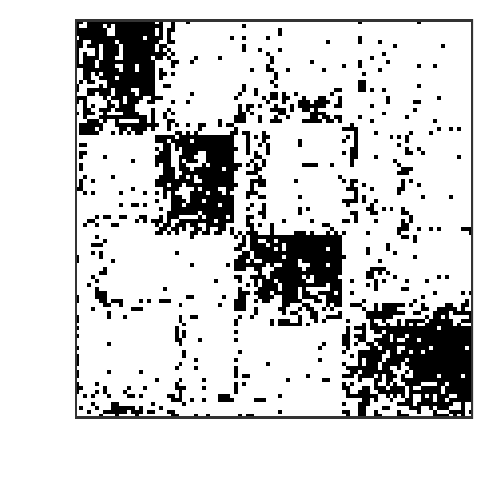}
}
\caption{Synthetic $100 \times 100$ matrices drawn from the three generative models with $K=4$. Matrices on the left are generated with $\alpha_k = \beta_k = \gamma_k = \eta_k = 1$. Matrices on the right are generated with $\alpha_k = \beta_k = \gamma_k = \eta_k = 0.1$. For better visualization, rows and columns are re-ordered according to complete linkage clustering \citep{Sorensen1948} using the $\texttt{hclust}$ function in R \citep{Rcran}.}
\label{fig:synthetics}
\end{figure}

Note that each element $w_{fk}$ and $h_{kn}$ can be interpreted as a probability. We can either merely impose that the elements of a row $\mathbf{w}_f$ or a column $\mathbf{h}_n$ lie between $0$ and $1$, or, more strongly, that they sum up to one. This implies a difference in modeling. On the one hand, imposing that the elements lie between $0$ and $1$ induce non-exclusive components. On the other hand, the sum-to-one constraint induces exclusive components, i.e., the more likely a component is, the less likely are the others. Fig.~\ref{fig:synthetics} displays simulated matrices generated from each of the models.

The first two models, \texttt{Beta-Dir} and \texttt{Dir-Beta}, are symmetric. Indeed, estimating $\mathbf{W}$ (resp., $\mathbf{H}$) in one model is equivalent to estimating $\mathbf{H}$ (resp., $\mathbf{W}$) in the other model after transposing the matrix $\mathbf{V}$. As such, in the rest of the paper, we will only consider the \texttt{Beta-Dir} model and the \texttt{Dir-Dir} model.

The Aspect Bernoulli model of \cite{Bingham2009} is built over Eq.~\eqref{eq:bern-nolink}, and considers that the factors $\mathbf{W}$ and $\mathbf{H}$ are deterministic parameters which satisfy the constraint (c2). The factors are estimated by maximum likelihood with EM. The binary ICA of \cite{Kaban2008} corresponds to the \texttt{Beta-Dir} model, and inference is performed with VB. In the following sections, we present inference methods for the posterior distributions of $\mathbf{W}$ and $\mathbf{H}$ in the \texttt{Beta-Dir} and \texttt{Dir-Dir} models. Because these distributions are intractable, we propose novel collapsed Gibbs sampling and collapsed variational inference strategies. We also derive a nonparametric approximation where the number of latent dimensions $K$ does not need to be fixed a priori.

\new{
\subsection{\ced{Connections with latent Dirichlet allocation (LDA)}.}
The proposed models have connection with topic models, in particular with LDA \citep{blei2003latent}, described by
\begin{align}
    \mathbf{h}_n & \sim \text{Dirichlet}(\bs{\gamma}) \\
    \mathbf{v}_n|\mathbf{h}_n & \sim {\text{Multinomial} \left( L_n, \ve{W} \ve{h}_n \right)}.
\end{align}
Here, $\mathbf{V}$ is the so-called ``bag-of-words'' representation of a corpus of documents, i.e., $v_{fn}$ represents the number of occurrences of word $f$ in document $n$. $L_n$ is the total number of words in document $n$, i.e., $L_n = \sum_f v_{fn}$. The columns of $\mathbf{W}$ are assumed to sum to 1, such that $w_{fk}$ represents the probability of word $f$ in topic $k$, and $h_{kn}$ represents the probability of topic $k$ in document $n$. In standard LDA, no prior distribution is assumed on $\mathbf{W}$, though it is common practice to assume Dirichlet distributions column-wise. The observation model is multinomial, with a total ``budget'' of $L_n$ words to distribute in each document $n$. In contrast, the binary data models presented in this article are based on independent Bernoulli observations, and cannot be recast as multinomial. An LDA-like model may arise if we assume that each column of $\mathbf{V}$ contains only one 1, and the rest are zeros (i.e. $L_n = 1$ for all $n$). This setting is however not considered in the article.}



\subsection{\ced{Bayesian inference}}
\ced{
In this paper, we opt for Bayesian inference of the latent factors $\ve{W}$ and $\ve{H}$. This means that we seek to characterize the posterior distribution $p(\ve{W},\ve{H}|\ve{V})$. The posterior in the considered models is not available in closed form and we will resort to numerical approximations (Markov Chain Monte Carlo sampling, variational inference). Another possible route is to seek point estimates through either constrained maximum likelihood (ML) or maximum a posteriori (MAP) estimation. For example, given Eq.~\eqref{eq:bern-nolink} and the set of constraints (c1), maximum likelihood estimation writes
\begin{align}
        \min_{\mathbf{W},\mathbf{H}} - \log p(\mathbf{V}|\mathbf{W} \mathbf{H}) = \sum_{f,n} v_{fn}\log([\ve{W}\ve{H}]_{fn}) + (1-v_{fn})\log(1-[\ve{W}\ve{H}]_{fn}) \nonumber \\
        \text{subject to} \quad (1)\ w_{fk}, h_{kn} \ge 0, \quad (2)\ \sum_{k} w_{fk}=1, \quad (3)\ h_{kn} \in [0,1].
\end{align}
Such optimization problems may be tackled with, e.g., majorization-minimization \citep{Bingham2009} or proximal gradient descent \citep{Udell16}. Though it can be computationally more costly, we favored Bayesian inference for the following reasons. Characterizing the full posterior provides a measure of uncertainty over the latent parameters and over predicted values of $\ve{V}$. As we will show later, it allows in turn to infer the rank $K$ of the factorization, a very desirable property that is more difficult to obtain with point estimation methods. Finally, Bayesian inference is rather customary in topic models such as LDA or Discrete Component Analysis \citep{bunt06}, and our work intends to follow similar principles.} 

\section{Inference in the \texttt{Beta-Dir} model}
In this section, we derive a collapsed Gibbs sampler \citep{Liu1994} for the \texttt{Beta-Dir} model. First, we will augment the model with latent indicator variables $\bs{Z}$ so that it becomes conjugate. Then the collapsed Gibbs sampler consists in marginalizing out the factors $\bs{W}$ and $\bs{H}$, thus running a Gibbs sampler over the indicator variables $\bs{Z}$ only. The interest of collapsed Gibbs sampling is that it offers improved mixing properties of the Markov chains, i.e., better exploration of the parameter space, thanks to the reduced dimensionality.  We use a superscript, as in $x^{(j)}$, to indicate the $j$-th sample of a chain (after burn-in). After sampling, given a collection of samples $\bs{Z}^{(j)}$ from the posterior, we will be able to directly sample from the posteriors of interest $p(\bs{W} | \bs{V})$ and $p(\bs{H} | \bs{V})$.

\subsection{Augmented model}\label{subsec:betadir_augmented}
\begin{figure}[t]
\centering
\subcaptionbox{Partially augmented \texttt{Beta-Dir} model}{
  \resizebox{0.8\columnwidth}{!}{\begin{tikzpicture}
\newcommand{\mynodedistance}{5mm}
\tikzstyle{observed}=[circle, minimum size = 8mm, thick, draw =black!80, fill = black!10, node distance = \mynodedistance]
\tikzstyle{latent}=[circle, minimum size = 8mm, thick, draw =black!80, node distance = \mynodedistance]
\tikzstyle{latent_invisible}=[circle, minimum size = 8mm,  node distance = \mynodedistance]
\tikzstyle{parameter}=[circle, minimum size = 8mm, thick, node distance = \mynodedistance]
\tikzstyle{connect}=[-latex, thick]
\tikzstyle{plate}=[rectangle, draw=black!100, rounded corners=0.25cm]
\tikzstyle{plate_invisible}=[rectangle, rounded corners=0.25cm]

  \node[parameter] (gamma) {$\bs{\gamma}$ };
  \node[latent]    (w) [right=1cm of gamma, xshift=0] {$\bs{w}_f$ };
  \node[latent]    (z) [right=of w] {$\bs{z}_{fn}$};
  \node[observed]  (v) [right=of z, xshift=0cm] { $v_{fn}$ };
  \node[latent_invisible]   (c) [right=of v, xshift=0cm] {};
  \node[latent]    (h) [right=of c] {$h_{kn}$};
  \node[parameter] (alpha) [right=of h,  xshift=1cm, yshift=0.5cm] {$\alpha_k$};
  \node[parameter] (beta)  [right=of h,  xshift=1cm, yshift=-0.5cm] {$\beta_k$};
  \path (gamma) edge [connect] (w)
        (w) edge [connect] (z)
		(z) edge [connect] (v)
		(h) edge [connect] (v)
		(alpha) edge [connect] (h)
		(beta) edge [connect] (h);
  \node[plate, 	   fit= (w)(z)(v), inner sep=4.6mm, xshift=-1mm] {}; 
  \node[rectangle, fit= (w), 		  inner sep=0mm, label=below left:$F$, xshift=1mm] {}; 
  \node[plate, 	   fit= (z)(v)(c)(h), inner sep=5.4mm,  minimum width=6.75cm, xshift=5mm] {}; 
  \node[rectangle, fit= (h), 		  inner sep=0mm, label=below right:{$N$}, xshift=7mm] {}; 
  \node[plate, 	   fit= (h), 	      inner sep=4mm, minimum width=2.1cm, xshift=0mm] {}; 
  \node[rectangle, fit= (h), 		  inner sep=0mm, label=below right:{$K$}, xshift=-0.5mm, yshift=0.5mm] {}; 

\end{tikzpicture}}
}
\subcaptionbox{Fully augmented \texttt{Dir-Dir} model}{
  \resizebox{0.8\columnwidth}{!}{\begin{tikzpicture}
\newcommand{\mynodedistance}{5mm}

\tikzstyle{observed}=[circle, minimum size = 8mm, thick, draw =black!80, fill = black!10, node distance = \mynodedistance]
\tikzstyle{latent}=[circle, minimum size = 8mm, thick, draw =black!80, node distance = \mynodedistance]
\tikzstyle{latent_invisible}=[circle, minimum size = 8mm,  node distance = \mynodedistance]
\tikzstyle{parameter}=[circle, minimum size = 8mm, thick, node distance = \mynodedistance]
\tikzstyle{connect}=[-latex, thick]
\tikzstyle{plate}=[rectangle, draw=black!100, rounded corners=0.25cm]
\tikzstyle{plate_invisible}=[rectangle, rounded corners=0.25cm]

  \node[parameter] (gamma) {$\bs{\gamma}$ };
  \node[latent]    (w) [right=1cm of gamma, xshift=0] {$\bs{w}_f$ };
  \node[latent]    (z) [right=of w] {$\bs{z}_{fn}$};
  \node[observed]  (v) [right=of z, xshift=0cm] { $v_{fn}$ };
  \node[latent]    (c) [right=of v, xshift=0cm] {$\bs{c}_{fn}$};
  \node[latent]    (h) [right=of c] {$\bs{h}_{n}$};
  \node[parameter] (eta) [right=1cmof h] {$\bs{\eta}$};
  \path (gamma) edge [connect] (w)
        (w) edge [connect] (z)
		(z) edge [connect] (v)
		(c) edge [connect] (v)
		(h) edge [connect] (c)
		(eta) edge [connect] (h);  
  \node[plate, 	   fit=    (w)(z)(v)(c), inner sep=4.6mm, xshift=-1mm] {}; 
  \node[plate, 	   fit= 	  (z)(v)(c)(h), inner sep=3.8mm, xshift=1mm] {}; 
  \node[rectangle, fit= (w), 		  inner sep=0mm, label=below left:$F$, xshift=1mm] {}; 
  \node[rectangle, fit= (h), 		  inner sep=0mm, label=below right:$N$, xshift=-1mm, yshift=1mm] {}; 

\end{tikzpicture}}
}
\caption{Augmented models}
\label{fig:augmentations}
\end{figure}
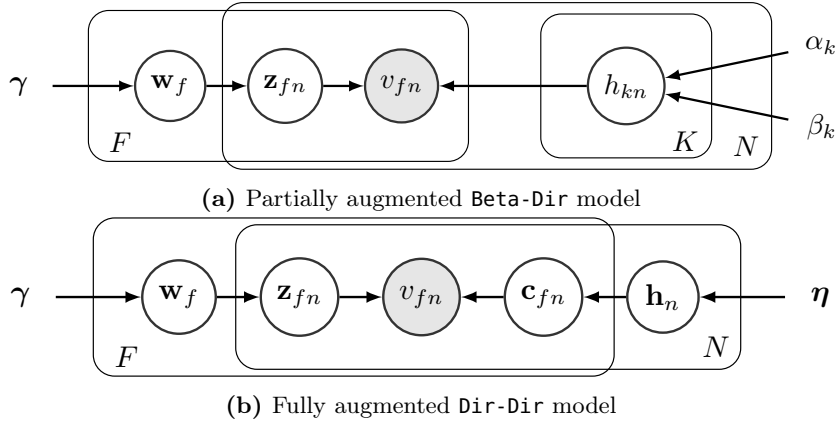

We can augment the \texttt{Beta-Dir} model with indicator variables $\bs{z}_{fn}$, that contain component assignments from the Dirichlet factor, as shown in \cite{Bingham2009}.  More precisely, $\bs{z}_{fn}$ is a vector of dimension $K$ with elements $z_{fkn} \in \{ 0, 1\}$ such that only one element equals to one and all the others equal to zero. In other words, $\bs{z}_{fn} \in \{ \bs{e}_{1}, \ldots, \bs{e}_{K} \}$, where $\bs{e}_k$ is the $k$-th canonical vector of $\mathbb{R}^K$. The augmented model is a mixture model described by:\footnote{\new{Some readers may be more accustomed to the alternative notation where the ``one-hot" variable $\bs{z}_{fn}$ is replaced by an integer-valued index $z_{fn} \in \{1,\ldots,K \}$. In this case, the Bernoulli parameter in Eq.~\eqref{eq:betadir-aug4} becomes $h_{z_{fn}n}$.}}
\begin{align}
    h_{kn} & \sim \text{Beta}(\alpha_k, \beta_k) \label{eq:betadir-aug1}\\
    \bs{w}_{f} & \sim \text{Dirichlet}(\bs{\gamma}) \label{eq:betadir-aug2} \\
    \bs{z}_{fn} | \bs{w}_f & \sim \text{Discrete}(\bs{w}_f) \label{eq:betadir-aug3}\\
    v_{fn} | \bs{h}_{n}, \bs{z}_{fn} & \sim \text{Bernoulli}\left(\prod_k h_{kn}^{z_{fkn}}\right). \label{eq:betadir-aug4}
\end{align}
{Indeed, marginalizing $\bs{z}_{fn}$ from Eqs.~\eqref{eq:betadir-aug3}-\eqref{eq:betadir-aug4} leads to Eq.~\eqref{eq:bern-nolink} as shown next. From Eqs.~\eqref{eq:betadir-aug3}-\eqref{eq:betadir-aug4} we have 
\begin{align}
p(v_{fn} | \ve{w}_{f}, \bs{h}_{n})& = \sum_{k} p(\bs{z}_{fn} = \bs{e}_{k} |\bs{w}_{f}) \, \text{Bernoulli}(v_{fn}|h_{kn}, z_{fkn}) \\
&= \sum_{k} w_{fk} \, h_{kn}^{v_{fn}} (1-h_{kn})^{1-v_{fn}},
\end{align}
and thus $p(v_{fn} = 1  | \ve{w}_{f},\bs{h}_{n}) = [\bs{W} \bs{H}]_{fn}$ and $p(v_{fn} = 0  | \ve{w}_{f},\bs{h}_{n}) = 1-[\bs{W} \bs{H}]_{fn}$, i.e., $v_{fn}$ has the marginal distribution given by Eq.~\eqref{eq:bern-nolink}.} A graphical representation of the augmented model is shown in Fig.~\ref{fig:augmentations}-(a). Let us think of a recommender system application, where columns of $\bs{V}$ are users, and rows are items. An interpretation of the above model is the following.
Each item $f$ is characterized by a probability over topics, $\bs{w}_f$. Then, for each user-item pair, a topic $k$ (indicated by $\bs{z}_{fn}$) is activated with probability $w_{fk}$, and the probability that the user $n$ consumes this item ($v_{fn}  = 1$) is $h_{kn}$.

Denoting by $\mathbf{Z}$ the $F \times K \times N$ tensor with entries $z_{fkn}$, the joint probability in the augmented model is given by:
\begin{align}
&p(\bs{V}, \bs{Z}, \bs{W}, \bs{H}) =
p(\bs{W})p(\bs{Z} | \bs{W})
p(\bs{H})p(\bs{V} | \bs{H}, \bs{Z})=
\notag\\
& 
\left[
\prod_{f=1}^F 
\left(
p(\bs{w}_f)
\prod_{n=1}^N
p(\bs{z}_{fn} | \bs{w}_{f})
\right)
\right]
\left[
\prod_{n=1}^N
\left(
\prod_{k=1}^K 
p(h_{kn})
\prod_{f=1}^F
p(v_{fn} | \bs{h}_n, \bs{z}_{fn}) 
\right)
\right].
\label{eq:joint_betadir}
\end{align}

\subsection{Collapsed Gibbs sampler}\label{subsec:betadir_gibbs}

Thanks to the previous augmentation, and exploiting conjugacy, we can now marginalize out $\bs{W}$ and $\bs{H}$ from Eq.~\eqref{eq:joint_betadir}. The marginalized distribution has the following structure:
\begin{align}
&p(\mathbf{V}, \mathbf{Z}) =
\notag\\
&\prod_f 
\overbrace{
\int
p(\bs{w}_{f})
\prod_{n}
p(\bs{z}_{fn} | \bs{w}_f) \,d\mathbf{w}_f
}^{p(\underline{\bs{Z}}_{f})
}
\prod_n
\overbrace{
\int
\prod_{k} 
p(h_{kn})
\prod_{f}
p(v_{fn} | \bs{h}_n, \bs{z}_{fn})
\, d\ve{h}_{n}
}^{p(\bs{v}_n | \bs{Z}_n)
},
\label{eq:marginalized_BetaDir}
\end{align}
where $\underline{\mathbf{Z}}_f$ denotes the $K \times N$ matrix with entries $\{ z_{fkn} \}_{kn}$, $\mathbf{Z}_n$ denotes the $F \times K$ matrix with entries $\{ z_{fkn} \}_{fk}$. Let us define the following four variables that act as counters: 
\begin{align*}
L_{fk} &= \sum_{n} z_{fkn},
& M_{kn} &= \sum_{f} z_{fkn},
\\
A_{kn} &= \sum_{f} z_{fkn}v_{fn},
& B_{kn} &= \sum_{f} z_{fkn}\bar{v}_{fn}.
\end{align*}
where $\bar{v}_{fn} = 1-v_{fn}$. $A_{kn}$ and $B_{kn}$ count how many times the component $k$ is associated to a ``positive'' observation ($v_{fn}=1$) and to a ``negative'' observation ($\bar{v}_{fn}  =1$) in the $n$-th sample. 
\new{Then we have the following expressions (derivations are available in Appendix \ref{sec:appendix_BetaDir_marginals}):}
\begin{align}
p(\underline{\bs{Z}}_f ) 
&=
\frac{\Gamma(\sum_k \gamma_k)}{\prod_k \Gamma(\gamma_k)}
\frac{\prod_k\Gamma(\gamma_k + L_{fk})}{\Gamma(\sum_k \gamma_k + N)}
\label{eq:marginalized_BetaDir_Z}\\
p(\bs{v}_n | \bs{Z}_n)
&=
\prod_{k}
\frac{\Gamma(\alpha_k + \beta_k)}{\Gamma(\alpha_k)\Gamma(\beta_k)}
\frac{\Gamma(\alpha_k + A_{kn})
\Gamma(\beta_k + B_{kn})
}{\Gamma(\alpha_k + \beta_k + M_{kn})}.
\label{eq:marginalized_BetaDir_V}
\end{align}

\begin{figure}[t]
\subcaptionbox{Observed matrix}{
\resizebox{0.24\textwidth}{!}{
\begin{tikzpicture}[>=latex]
\newcommand{\mynodedistance}{5mm}
\tikzstyle{observed}=[circle, minimum size = 8mm, thick, draw =black!80, fill = black!10, node distance = \mynodedistance]
\tikzstyle{latent}=[circle, minimum size = 8mm, thick, draw =black!80, node distance = \mynodedistance]
\tikzstyle{latent_invisible}=[circle, minimum size = 8mm,  node distance = \mynodedistance]
\tikzstyle{parameter}=[circle, minimum size = 8mm, thick, node distance = \mynodedistance]
\tikzstyle{connect}=[-latex, thick]
\tikzstyle{plate}=[rectangle, draw=black!100, rounded corners=0.25cm]
\tikzstyle{plate_invisible}=[rectangle, rounded corners=0.25cm]

\matrix[row sep=1*\myunit, column sep=2*\myunit] (V)%
		     [matrix of math nodes,%
             nodes = {node style ge},%
             left delimiter  = (,%
             right delimiter = )] at (0,0)
{%
  \node[node style one] {};&
  \node[node style zero] {};&
  \node[node style zero] {};&
  \node[node style one] {};&
  \node[node style zero] {};&
  \node[node style zero] {};&  
  \node[node style zero] {};  \\
  \node[node style one] {};&
  \node[node style zero] {};&
  \node[node style one] {};&
  \node[node style zero] {};&
  \node[node style one] {};&
  \node[node style zero] {};&  
  \node[node style one] {};  \\
  \node[node style one] {};&
  \node[node style zero] {};&
  \node[node style zero] {};&
  \node[node style one] {};&
  \node[node style zero] {};&
  \node[node style zero] {};&  
  \node[node style one] {};  \\
  \node[node style zero] {};&
  \node[node style one] {};&
  \node[node style zero] {};&
  \node[node style zero] {};&
  \node[node style zero] {};&
  \node[node style zero] {};&  
  \node[node style one] {};  \\
  \node[node style one] {};&
  \node[node style zero] {};&
  \node[node style zero] {};&
  \node[node style one] {};&
  \node[node style zero] {};&
  \node[node style one] {};&  
  \node[node style one] {};  \\
};
\node [below=10pt] at (V.south) {{$\bs{V}$}};

\end{tikzpicture}}
}%
\subcaptionbox{\texttt{Beta-Dir}}{
\resizebox{0.24\textwidth}{!}{
\begin{tikzpicture}[>=latex]
\newcommand{\mynodedistance}{5mm}
\tikzstyle{observed}=[circle, minimum size = 8mm, thick, draw =black!80, fill = black!10, node distance = \mynodedistance]
\tikzstyle{latent}=[circle, minimum size = 8mm, thick, draw =black!80, node distance = \mynodedistance]
\tikzstyle{latent_invisible}=[circle, minimum size = 8mm,  node distance = \mynodedistance]
\tikzstyle{parameter}=[circle, minimum size = 8mm, thick, node distance = \mynodedistance]
\tikzstyle{connect}=[-latex, thick]
\tikzstyle{plate}=[rectangle, draw=black!100, rounded corners=0.25cm]
\tikzstyle{plate_invisible}=[rectangle, rounded corners=0.25cm]


\matrix[row sep=1*\myunit, column sep=2*\myunit] (Z) [matrix of math nodes,%
             nodes = {node style ge},%
             left delimiter  = (,%
             right delimiter = )] at (5,0)
{%
  \node[node style na] {};&
  \node[node style blue, ultra thick] {};&
  \node[node style green, ultra thick] {};&
  \node[node style red, ultra thick] {};&
  \node[node style green, ultra thick] {};&
  \node[node style blue, ultra thick] {};&  
  \node[node style red, ultra thick] {};  \\
  \node[node style green, ultra thick] {};&
  \node[node style green] {};&
  \node[node style blue] {};&
  \node[node style blue] {};&
  \node[node style red] {};&
  \node[node style red] {};&  
  \node[node style blue] {};  \\
  \node[node style blue, ultra thick] {};&
  \node[node style blue] {};&
  \node[node style blue] {};&
  \node[node style blue] {};&
  \node[node style green] {};&
  \node[node style green] {};&  
  \node[node style blue] {};  \\
  \node[node style blue] {};&
  \node[node style blue] {};&
  \node[node style blue] {};&
  \node[node style blue] {};&
  \node[node style red] {};&
  \node[node style blue] {};&  
  \node[node style blue] {};  \\
  \node[node style green, ultra thick] {};&
  \node[node style red] {};&
  \node[node style green] {};&
  \node[node style green] {};&
  \node[node style green] {};&
  \node[node style red] {};&  
  \node[node style red] {};  \\
};
\node [below=10pt] at (Z.south) {{$\bs{Z}$}};

\end{tikzpicture}}
}%
\subcaptionbox{\texttt{Dir-Dir}}{
\resizebox{0.47\textwidth}{!}{
\begin{tikzpicture}[>=latex]

\matrix[row sep=1*\myunit, column sep=2*\myunit] (Z) [matrix of math nodes,%
             nodes = {node style ge},%
             left delimiter  = (,%
             right delimiter = )] at (5,0)
{%
  \node[node style na] {};&
  \node[node style blue, ultra thick] {};&
  \node[node style green, ultra thick] {};&
  \node[node style red, ultra thick] {};&
  \node[node style green,ultra thick] {};&
  \node[node style blue, ultra thick] {};&  
  \node[node style red, ultra thick] {};  \\
  \node[node style green] {};&
  \node[node style green] {};&
  \node[node style blue] {};&
  \node[node style blue] {};&
  \node[node style red] {};&
  \node[node style red] {};&  
  \node[node style blue] {};  \\
  \node[node style blue] {};&
  \node[node style blue] {};&
  \node[node style blue] {};&
  \node[node style blue] {};&
  \node[node style green] {};&
  \node[node style green] {};&  
  \node[node style blue] {};  \\
  \node[node style blue] {};&
  \node[node style blue] {};&
  \node[node style blue] {};&
  \node[node style blue] {};&
  \node[node style red] {};&
  \node[node style blue] {};&  
  \node[node style blue] {};  \\
  \node[node style green] {};&
  \node[node style red] {};&
  \node[node style green] {};&
  \node[node style green] {};&
  \node[node style green] {};&
  \node[node style red] {};&  
  \node[node style red] {};  \\
};
\node [below=10pt] at (Z.south) {{$\bs{Z}$}};

\matrix[row sep=1*\myunit, column sep=2*\myunit] (Z) [matrix of math nodes,%
             nodes = {node style ge},%
             left delimiter  = (,%
             right delimiter = )] at (9.3,0)
{%
  \node[node style na, ] {};&
  \node[node style blue] {};&
  \node[node style blue] {};&
  \node[node style red] {};&
  \node[node style red] {};&
  \node[node style green] {};&  
  \node[node style blue] {};  \\
  \node[node style green, ultra thick] {};&
  \node[node style red] {};&
  \node[node style red] {};&
  \node[node style blue] {};&
  \node[node style green] {};&
  \node[node style red] {};&  
  \node[node style blue] {};  \\
  \node[node style green, ultra thick] {};&
  \node[node style red] {};&
  \node[node style red] {};&
  \node[node style blue] {};&
  \node[node style green] {};&
  \node[node style green] {};&  
  \node[node style blue] {};  \\
  \node[node style blue, ultra thick] {};&
  \node[node style green] {};&
  \node[node style blue] {};&
  \node[node style green] {};&
  \node[node style red] {};&
  \node[node style blue] {};&  
  \node[node style red] {};  \\
  \node[node style green, ultra thick] {};&
  \node[node style red] {};&
  \node[node style green] {};&
  \node[node style green] {};&
  \node[node style green] {};&
  \node[node style red] {};&  
  \node[node style green] {};  \\
};
\node [below=10pt] at (Z.south) {{$\bs{C}$}};

\end{tikzpicture}}
}
\caption{Illustration of the Gibbs samplers. Colors represent component assignments {(a value of $k$)}. When sampling an element (dashed circle) the probability of each component depends on the number of elements \ced{with thick circles} in the same row or column that are currently assigned to that component.}
\label{fig:gibbs}
\end{figure}
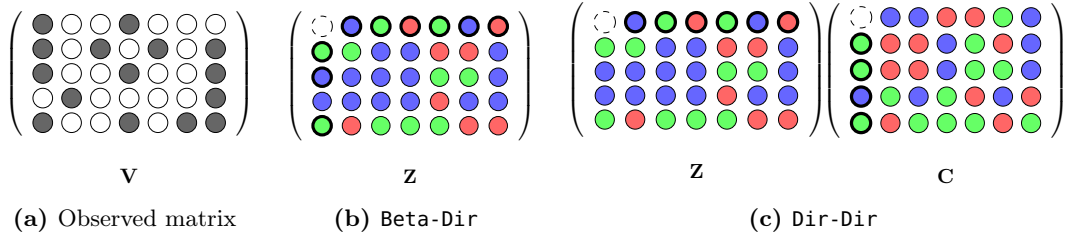

The posterior of the indicator variables $p(\bs{Z}|\bs{V})$ is not available in closed form and the proposed collapsed Gibbs sampler consists in iteratively sampling each vector $\bs{z}_{fn}$ given the current value of the other indicator vectors. Let $L_{fk}^{\neg fn}$, $M_{kn}^{\neg fn}$, $A_{kn}^{\neg fn}$, $B_{kn}^{\neg fn}$ be the state of the counters when the tube $(f,n)$ of the tensor $\bs{Z}$ is left out of the sums:
\begin{align*}
L_{fk}^{\neg fn} &= L_{fk} - z_{fkn},
& M_{kn}^{\neg fn} &= M_{kn} - z_{fkn},
\\
A_{kn}^{\neg fn} &= A_{kn} - z_{fkn}v_{fn},
& B_{kn}^{\neg fn} &= B_{kn} - z_{fkn}\bar{v}_{fn}.
\end{align*}
In Appendix~\ref{sec:appendix_BetaDir_conditionals} we show that the conditional posterior of $\bs{z}_{fn}$ given the remaining variables $ \bs{Z}_{\neg fn}$ is given by:
\begin{align}
p(\bs{z}_{fn} | \bs{Z}_{\neg fn},\bs{V}) \propto
\prod_k
\left[
(\gamma_k + L_{fk}^{\neg fn} ) 
\frac{(\alpha_k + A_{kn}^{\neg fn})^{v_{fn}}
(\beta_k + B_{kn}^{\neg fn})^{\bar{v}_{fn}}
}{\alpha_k + \beta_k + M_{kn}^{\neg fn}}
\right]^{z_{fkn}}.
\label{eq:conditional_z_DirBer}
\end{align}
This expression needs to be normalized to ensure a valid probability distribution. This can be easily done by computing the right-hand side of Eq.~\eqref{eq:conditional_z_DirBer} for every of the $K$ possible values of $\bs{z}_{fn}$ and normalizing by the sum. Eq.~\eqref{eq:conditional_z_DirBer} shows that the probability of choosing a component $k$ depends on the number of elements already assigned to that component. More precisely, it depends on the one hand on the number of elements assigned to component $k$ in column $n$. On the other hand, it also depends on the proportion of elements in row $f$ assigned to component $k$ that explain ones (if $v_{fn}=1$) or zeros (if $v_{fn}=0$) in $\bs{V}$ in the total number of elements associated to $k$ in that row (see Figure~\ref{fig:gibbs}). The parameters $\gamma_k, \alpha_k, \beta_k$ act as pseudo-counts: they give \textit{a priori} belief about how many elements are assigned to each component.

Our collapsed Gibbs sampling is summarized in Alg.~\eqref{algo:gibbsBetaDir}. Note that, although Alg.~\eqref{algo:gibbsBetaDir} does not explicitly include it, we must draw samples during an initial burn-in phase (as required by any MCMC method) before collecting the last $J$ samples, after the chain has converged to the stationary distribution. 
{Note also that the algorithm can readily deal with incomplete matrices by simply skipping missing entries (i.e., the loop over $f$ and $n$ only runs over available entries).} \\

\begin{algorithm}[t]
 \SetKwInput{Input}{Input}
 \SetKwInput{Output}{Output}
 \SetKwInput{Parameters}{Parameters}
 \SetKwInput{Initialize}{Initialize}
 \Input{Observed matrix $\bs{V}\in\{0,1\}^{F \times N}$}
 \Parameters{$\bs{\alpha}, \bs{\beta}, \bs{\gamma}$}
 \Output{Samples $\bs{Z}^{(1)},...,\bs{Z}^{(J)}$}
 \Initialize{Random initialization of $\bs{Z}$}
 \For{$j=1$ \KwTo $J$ }{
 \For{$f=1$ \KwTo $F$ }{
 \For{$n=1$ \KwTo $N$ }{
 	\If{$v_{fn}$ not missing}{
   Sample $\bs{z}_{fn}^{(j)} \sim p(\bs{z}_{fn} | \bs{Z}_{\neg fn}, \bs{V})$ (Eq.\eqref{eq:conditional_z_DirBer}) 
   }
 }
 }
 }
 \caption{Collapsed Gibbs sampler for \texttt{Beta-Dir}}
 \label{algo:gibbsBetaDir}
\end{algorithm}

\noindent \textbf{Latent factors posteriors.} Thanks to conjugacy, the conditional posteriors of $\mathbf{W}$ and $\mathbf{H}$ given $\bs{Z}$ and $\bs{V}$ are given by:
\begin{align}
\bs{w}_f | \underline{\bs{Z}}_f &\sim \text{Dirichlet}(\bs{\gamma} + \sum_n \bs{z}_{fn}) \label{eqn:Pw}
\\ 
h_{kn} | \bs{Z}_n, \bs{v}_n  &\sim \text{Beta}(\alpha_k + A_{kn},~\beta_k +B_{kn}). \label{eqn:Ph}
\end{align}
The conditional posterior expectations are given by:
\begin{align}
\mathbb{E}_{\bs{w}_f}[\bs{w}_f | \underline{\bs{Z}}_f] &= 
\frac{\bs{\gamma} + \sum_n \bs{z}_{fn}}{\sum_k \gamma_k + \sum_k\sum_n z_{fkn}}
=
\frac{\bs{\gamma} + \sum_n \bs{z}_{fn}}{\sum_k \gamma_k + N}
\label{eqn:Ew} \\
\mathbb{E}_{h_{kn}}[h_{kn} | \bs{Z}_n, \bs{v}_n] & = 
\frac{\alpha_k + A_{kn}}{\alpha_k + A_{kn} + \beta_k + B_{kn}}
=
\frac{\alpha_k + \sum_f z_{fkn}v_{fn}}{\alpha_k +  \beta_k + \sum_f z_{fkn}},
\label{eqn:Eh}
\end{align}
where we used the {equalities $\sum_{kn} z_{fkn} = N $} and $A_{kn} + B_{kn} = \sum_f z_{fkn}$.
Using the law of total expectation, i.e., $\mathbb{E}[X]~=~\mathbb{E}_Y[\mathbb{E}_X[X | Y]]$, and given a set a samples $\bs{Z}^{(j)}$, it follows that the marginal posterior expectations of the latent factors can be computed as:
\begin{align}
\mathbb{E}_{\bs{w}_f}[\bs{w}_f | \bs{V}] &= 
\mathbb{E}_{\bs{Z}}[\mathbb{E}_{\bs{w}_f}[\bs{w}_f | \bs{V}, \bs{Z}]]
=
\mathbb{E}_{\underline{\bs{Z}}_f}[\mathbb{E}_{\bs{w}_f}[\bs{w}_f |  \underline{\bs{Z}}_f]]
\approx \frac{\bs{\gamma} + \frac{1}{J} \sum_j \sum_n \mathbf{z}_{fn}^{(j)}}
{\sum_k \gamma_k + N}
\\
\mathbb{E}_{h_{kn}}[h_{kn} | \bs{V}] &= 
\mathbb{E}_{\bs{Z}_{n}}[\mathbb{E}_{h_{kn}}[h_{kn}| \bs{Z}_{n},\bs{v}_{n}]]
\approx 
\frac{1}{J}
\sum_j
\frac{\alpha_k + \sum_f z_{fkn}^{(j)}v_{fn}}{\alpha_k + \beta_k + \sum_f z_{fkn}^{(j)}}.
\end{align}

\noindent \textbf{Prediction.} The predictive posterior distribution of an unseen data sample $v_{fn}^*$ given the available data $\bs{V}$ is given by
\begin{align}
p(v_{fn}^*|\bs{V}) &= \int p(v_{fn}^* | \bs{w}_f, \bs{h}_n)p(\bs{w}_f, \bs{h}_n | \bs{V})
 \,d\bs{w}_f d\bs{h}_n
\notag\\
&=
\mathbb{E}[\bs{w}_f \bs{h}_n | \bs{V}]^{v_{fn}^*}
(1-\mathbb{E}[\bs{w}_f \bs{h}_n | \bs{V}])^{1-v_{fn}^*}.
\label{eq:predictive_posterior}
\end{align}
Because the predictive posterior is a Bernoulli distribution, its expectation is given by $\mathbb{E}[\bs{w}_f \bs{h}_n | \bs{V}]$, which can be approximated using samples $\bs{W}^{(j)}$, $\bs{H}^{(j)}$ from the distributions given by Eqs.~\eqref{eqn:Pw}-\eqref{eqn:Ph} given $\bs{Z} = \bs{Z}^{(j)}$:
\begin{align}
\mathbb{E}[v_{fn}^* | \bs{V}] = \mathbb{E}[{\bs{w}_f \bs{h}_{n}} | \bs{V}] \approx \frac{1}{J}\sum_j  \bs{w}_f^{(j)} \bs{h}_n^{(j)}.
\label{eq:predictive_expectation}
\end{align}

\subsection{Collapsed variational inference}\label{subsec:betadir_vb}

Given the collapsed model of Eqs.~\eqref{eq:marginalized_BetaDir}-\eqref{eq:marginalized_BetaDir_V}
we may derive a mean-field Collapsed Variational Bayes algorithm (CVB) \citep{Teh2007a} by assuming that the posterior factorizes as
$
q(\bs{Z}) = \prod_{fn} q(\bs{z}_{fn})
$.
The key of CVB is that its free energy is a strictly better bound on the evidence than the free energy of the standard, i.e., uncollapsed, VB. 
We compute the CVB updates by applying the mean-field VB updates to the collapsed model:
\begin{align}
q(\bs{z}_{fn} | \bs{V}) \propto \exp\{ \mathbb{E}_{q(\bs{Z}_{\neg fn})}[\log p(\bs{V},\bs{Z})]\},
\end{align}
where the expectations are taken over the variational posterior. This leads us to 
\begin{align}
&q(\bs{z}_{fn} | \bs{V}) \notag\\
&\propto
\prod_k
\exp\left\lbrace
\mathbb{E}_q[\log (\gamma_k+ L_{fk}^{\neg fn})]
\frac{
\mathbb{E}_q[\log(\alpha_k + A_{kn}^{\neg fn})]^{v_{fn}}
\mathbb{E}_q[\log(\beta_k  + B_{kn}^{\neg fn})]^{\bar{v}_{fn}}
}{\mathbb{E}_q[\log M_{kn}^{\neg fn}]}
\right\rbrace.
\end{align}
The expectations of the form $\mathbb{E}_{q(z)}[\log(x + z)]$ are expensive to compute.
A simpler alternative is CVB0 \citep{Asuncion2009}, which uses a zero-order Taylor approximation $\mathbb{E}_q(z)[\log(x + z)] \approx \log(x + \mathbb{E}_{q(z)}[z])]$ and has been shown to give, in some cases, better inference results than CVB. Under the CVB0 version our update becomes
\begin{align}
q(\bs{z}_{fn} | \bs{V}) 
\propto 
\prod_k
(\gamma_k + \mathbb{E}_q[L_{fk}^{\neg fn}])
\frac
{(\alpha_k + \mathbb{E}_q[A_{kn}^{\neg fn}])^{v_{fn}}
(\beta_k + \mathbb{E}_q[B_{kn}^{\neg fn}])^{\bar{v}_{fn}}}
{\alpha_k + \beta_k +  \mathbb{E}_q[M_{kn}^{\neg fn}]},
\end{align}
which has a similar structure to the collapsed Gibbs sampler in Eq.~\eqref{eq:conditional_z_DirBer}. Overall, the collapsed VB algorithm has the same structure as the Gibbs sampler summarized in Alg.~\eqref{algo:gibbsBetaDir}. \ced{Note that when the data matrix is too large for batch processing, one can routinely resort to stochastic variational inference \citep{SVI}.} \\

\noindent \textbf{Latent factors posteriors.} The variational distributions of the factors can be obtained from the uncollapsed version:
\begin{align}
q(\bs{w}_f) &= \text{Dirichlet}(\bs{\gamma} + \sum_n\mathbb{E}_q[\bs{z}_{fn}])\\
q(h_{kn}) &= \text{Beta}(\alpha_k + \mathbb{E}_q[A_{kn}], \beta_k + \mathbb{E}_q[B_{kn}] )).
\end{align}
The Taylor approximation breaks the theoretical guarantees of their superiority over the ones given by uncollapsed VB. Still, they have been reported to work better than VB in practice. Another drawback of the approximation is the loss of convergence guarantees. Although we do not address this issue here, this has been recently addressed by \cite{Ishiguro2017a}, where an annealing strategy is used to gradually decrease the portion of the variational posterior changes. \\

\noindent{\bf Prediction.} The predictive posterior can be computed as in Eq.~\eqref{eq:predictive_posterior}, and its expectation is computed using the variational approximations of the factors, i.e.,
\begin{align}
\mathbb{E}[v_{fn}^* | \bs{V}] = \mathbb{E}[{\bs{w}_f \bs{h}_{n}} | \bs{V}] \approx  \mathbb{E}_{q(\bs{w}_f)}[\bs{w}_f]\,\mathbb{E}_{q(\bs{h}_{n})}[\bs{h}_n].
\label{eq:predictive_expectation_vb}
\end{align}

\subsection{Approximating infinite components}\label{subsec:betadir_inf}

Recall that in the augmented model, the component assignments $\bs{z}_{fn}$ have a Discrete distribution such that (Eqs.~\eqref{eq:betadir-aug2}-\eqref{eq:betadir-aug3})
\begin{align}
\bs{w}_{f} &\sim \text{Dirichlet}(\bs{\gamma})\notag\\
\bs{z}_{fn} | \bs{w}_{f} &\sim \text{Discrete}(\bs{w}_f).\notag
\end{align}
The variable $\bs{w}_f$ may be integrated out leading to the expression of $p(\underline{\bs{Z}}_f )$ given by Eq.\eqref{eq:marginalized_BetaDir_Z}. In Appendix~\ref{sec:appendix_BetaDir_conditionals}, we show that the prior conditionals are given by:
\begin{align}
p(\bs{z}_{fn} = \bs{e}_{k} |\bs{Z}_{\neg fn}) =
\frac{\gamma_{k} + L_{fk}^{\neg fn}}{\sum_{k} \gamma_{k} + N -1}.
\end{align}
{Let us assume from now that the Dirichlet prior parameters are such that $\gamma_k = \gamma/K$, where $\gamma$ is a fixed nonnegative scalar, so that:
\begin{align}
p(\bs{z}_{fn} = \bs{e}_{k} |\bs{Z}_{\neg fn}) =
\frac{\gamma/K + L_{fk}^{\neg fn}}{\gamma + N -1}. \label{eqn:CRP}
\end{align}
The conditional prior given by Eq.~\eqref{eqn:CRP} is reminiscent of the Chinese Restaurant process (CRP) \citep{Aldous1985,Anderson91,Pitman2002}. In the limit when $K \rightarrow \infty$, the probability of assigning $\bs{z}_{fn}$ to component $k$ is proportional to the number $L_{fk}^{\neg fn}$ of current assignments to that component. 
Let $K^+$ denote the current number of non-empty components (i.e., such that $L_{fk}^{\neg fn} >0$). Then the probability of choosing an empty component is
\begin{align}
p(\bs{z}_{fn}= \bs{e}_k |\bs{Z}_{\neg fn}, L_{fk}^{\neg fn}  = 0) 
=
\lim_{K \rightarrow \infty} (K - K^+)
\frac{\gamma/K}{\gamma + N -1} = 
\frac{\gamma}{\gamma + N -1}.
\end{align}
Note that the latter probability does not depend on $K$. In practice, we set $K$ to a large value and observed self-pruning of the number of components, hence achieving to automatic order selection, similar to \cite{Hoffman2010a}.  Implementing exact inference in the truly nonparametric model
\begin{align}
\underline{\bs{Z}}_{f} &\sim \text{CRP}({\gamma})\\
\quad \bs{v}_{n} | \bs{Z}_n &\sim p(\bs{v}_n | \bs{Z}_n)
\end{align}
is more challenging. This is because there is a CRP for each feature $f$, and some empty components may become unidentifiable in the limit. This is a known issue that could be addressed using for example a Chinese Restaurant Franchise process \citep{Teh2006} but is beyond the scope of this article.}

\section{Inference in the Dirichlet-Dirichlet model}\label{sec:dirdir}

The methodology to obtain a collapsed Gibbs sampler for the \texttt{Dir-Dir} model is similar to the approach followed for the \texttt{Beta-Dir} model.
\new{
It is possible to augment the model with the same auxiliary variable $\ve{z}_{fn}$ and compute the expression of $p(\bs{z}_{fn} | \bs{Z}_{\neg fn},\bs{V})$ in closed form. However, the expression of the conditional posterior, given in Appendix~\ref{sec:appendix_DirDir}, involves combinatorial computations and is infeasible in practice. As such, we propose an alternative Gibbs sampler that relies on a double augmentation, presented next.} Obtaining a variational collapsed algorithm for the \texttt{Dir-Dir} model is not straightforward, even using the double augmentation, and is left for future work.

\subsection{{Fully augmented model}}\label{subsec:dirdir_augmented}

Unlike the \texttt{Beta-Dir} model, the \texttt{Dir-Dir} model is not fully conjugate after a first augmentation. We {propose} a second augmentation with a new indicator variable $\bs{c}_{fn} \in \{ \bs{e}_{1}, \ldots, \bs{e}_{K} \}$, that plays a similar role to $\bs{z}_{fn}$ The \textit{fully} augmented version is:
\begin{align}
  \bs{h}_{n} &\sim \text{Dirichlet}(\bs{\eta})\\
  \bs{w}_{f} &\sim \text{Dirichlet}(\bs{\gamma})\\
   \bs{c}_{fn} | \bs{h}_n&\sim \text{Discrete}(\bs{h}_n)\\  
  \bs{z}_{fn} | \bs{w}_f &\sim \text{Discrete}(\bs{w}_f)\\
  v_{fn} &= {\sum_k c_{fkn} z_{fkn}}.
\end{align}
{To show that this is a valid augmentation}, note that $v_{fn}$ can only be nonzero (and equal to 1) if $\bs{c}_{fn} = \bs{z}_{fn}$. {Then}, the marginal probability of $v_{fn} =1$ is given by
\begin{align}
p(v_{fn} = 1 | \ve{w}_{f}, \ve{h}_{n}) &= \sum_{k} p(v_{fn} =1,  \ve{z}_{fn} = \ve{c}_{fn}= \ve{e}_{k}| \ve{w}_{f}, \ve{h}_{n}) \\
&= \sum_{k} p(\ve{z}_{fn} = \ve{e}_{k} | \ve{w}_{f}) p(\ve{c}_{fn} = \ve{e}_{k}| \ve{h}_{n}) \\
&= \sum_{k} w_{fk} h_{kn},
\end{align}
and we thus recover the Bernoulli model of Eq.~\eqref{eq:bern-nolink} as announced. Compared to the \texttt{Beta-Dir} model and using our recommender system analogy, this means that, in each user-item pair, the user also activates one topic, and then consumes the item if the user active topic is {equal to} the item active topic. The \texttt{Dir-Dir} model makes a stronger assumption than the \texttt{Beta-Dir} since the user can only activate one topic per item. A graphical representation of the fully augmented model is given in Fig.~\ref{fig:augmentations}-(b). In the following, we denote by $\bs{C}$ the $F \times K \times N$ tensor with entries $c_{fkn}$, and by $\bs{C}_n$ the $F \times K$ matrix with entries $\{c_{fkn}\}_{fk}$. 

\subsection{Collapsed Gibbs sampling}\label{subsec:dirdir_gibbs}

In this section we show that $\bs{W}$ and $\bs{H}$ can be marginalized from the joint probability of the fully augmented model and then propose a collapsed Gibbs sampler for $p(\bs{Z}, \bs{C} | \bs{V})$. The joint probability is given by:
\begin{align}
p(\bs{V}, \bs{Z}, \bs{C}) = \prod_{f,n}
p(v_{fn} | \bs{z}_{fn}, \bs{c}_{fn})
\prod_f p(\underline{\bs{Z}}_f)
\prod_n p(\bs{C}_n),
\end{align}
where
\begin{align}
p(\underline{\bs{Z}}_f) 
&=
\frac{\Gamma(\sum_k \gamma_k)}{\prod_k \Gamma(\gamma_k)}
\frac{\prod_k\Gamma(\gamma_k + \sum_n z_{fkn})}{\Gamma(\sum_k \gamma_k + N)}
\\
p(\bs{C}_n) 
&=
\frac{\Gamma (\sum_k \eta_k)}{\prod_k \Gamma(\eta_k)}
\frac{\prod_k\Gamma(\eta_k + \sum_f c_{fkn})}{\Gamma(\sum_k \eta_k + F)}\\
p(v_{fn} | \bs{z}_{fn}, \bs{c}_{fn}) &= {\delta(v_{fn} - \sum_{k} c_{fkn} z_{fkn})},
\end{align}
and where $\delta$ denotes the Dirac delta function. Following Section~\ref{subsec:betadir_gibbs} and Appendix~\ref{sec:appendix_BetaDir_conditionals}, the prior conditional are given by:
\begin{align}
p(\bs{z}_{fn} | \bs{Z}_{\neg fn}) &\propto \prod_k (\gamma_k + L_{fk}^{\neg fn})^{z_{fkn}}
\\
p(\bs{c}_{fn} | \bs{C}_{\neg fn}) &\propto \prod_k (\eta_k + {Q}_{kn}^{\neg fn})^{c_{fkn}}.
\end{align}
where $Q_{kn}^{\neg fn} = \sum_{f'\not=f} c_{f'kn}$ and $L_{fk}^{\neg fn}=\sum_{n'\not=n} z_{fkn'}$ is as before.
When $v_{fn}=1$, $\bs{c}_{fn}$ and $\bs{z}_{fn}$ must be assigned to the same component ($\bs{c}_{fn} = \bs{z}_{fn}$). To respect this constraint, we may sample them together from the posterior. Introducing the vector $\bs{x}_{fn}$ such that $\bs{x}_{fn} = \bs{z}_{fn} = \bs{c}_{fn}$, the conditional posterior is given by:
\begin{align}
p(\bs{x}_{fn} | \bs{Z}_{\neg fn}, \bs{C}_{\neg fn}, v_{fn}=1)
&\propto
\prod_k 
\left[(\gamma_k + L_{fk}^{\neg fn})
(\eta_k + Q_{kn}^{\neg fn})\right]^{{x_{fkn}}}.
\label{eq:conditional_zc_DirDir}
\end{align}
When $v_{fn}=0$, we can assign to one of the two auxiliary variables any component not currently assigned to the other auxiliary variable. The respective conditional posteriors are given by:
\begin{align}
p(\bs{z}_{fn}| \bs{Z}_{\neg fn}, \bs{c}_{fn}, v_{fn}=0)
&\propto
\prod_k
\left[
(\gamma_k + L_{fk}^{\neg fn})
(1-c_{fkn})\right]^{z_{fkn}}
\label{eq:conditional_z_DirDir}
\\
p(\bs{c}_{fn}| \bs{z}_{fn}, \bs{C}_{\neg fn}, v_{fn}=0)
&\propto
\prod_k
\left[
(\eta_k + Q_{kn}^{\neg fn})
(1-z_{fkn})
\right]^{c_{fkn}}.
\label{eq:conditional_c_DirDir}
\end{align}
A pseudo-code of the resulting Gibbs sampler is given in Alg.~\eqref{algo:gibbsDirDir}. As with the \texttt{Beta-Dir} model, we set $\gamma_{k} = \gamma/K$ with $K$ large to emulate a nonparametric setting (note that $\bs{\eta}$ does not need to depend on $K$ itself).\\

\begin{algorithm}[t]
 \SetKwInput{Input}{Input}
 \SetKwInput{Output}{Output}
 \SetKwInput{Parameters}{Parameters}
 \SetKwInput{Initialize}{Initialize}
 \Input{Observed matrix $\bs{V}\in\{0,1\}^{F \times N}$}
 \Parameters{$\bs{\gamma}, {\bs{\eta}}$}
 \Output{Samples $\bs{Z}^{(1)},...,\bs{Z}^{(J)}$, $\bs{C}^{(1)},...,\bs{C}^{(J)}$}
 \Initialize{Random initialization of $\bs{Z}$ and $\bs{C}$}
 \For{$j=1$ \KwTo $J$ }{
 \For{$f=1$ \KwTo $F$ }{
 \For{$n=1$ \KwTo $N$ }{
 	\If{$v_{fn}$ not missing}{
 \eIf{$v_{fn}=1$}{
   Sample $\bs{x} \sim p(\bs{x} | \bs{Z}_{\neg fn}, \bs{C}_{\neg fn}, \bs{V})$ (Eq.~\eqref{eq:conditional_zc_DirDir})\\ 
   $\bs{z}_{fn}^{(j)} = \bs{x}$\\
   $\bs{c}_{fn}^{(j)} = \bs{x}$\\
 }
 {
 Sample $\bs{z}_{fn}^{(j)} \sim p(z_{fn} | \bs{Z}_{\neg fn}, \bs{C}, \bs{V})$ (Eq.~\eqref{eq:conditional_z_DirDir})\\
 Sample $\bs{c}_{fn}^{(j)} \sim p(z_{fn} | \bs{Z}, \bs{C}_{\neg fn}, \bs{V})$ (Eq.~\eqref{eq:conditional_c_DirDir})
 }
 }
 }
 }
 }
 \caption{Collapsed Gibbs sampler for \texttt{Dir-Dir}}
 \label{algo:gibbsDirDir}
\end{algorithm}

\noindent \textbf{Latent factors posteriors and prediction.} The conditional posteriors of the latent factors given $\bs{Z}$ and $\bs{C}$ are given by:
\begin{align}
\bs{w}_f | \underline{\bs{Z}}_f &\sim \text{Dirichlet}(\bs{\gamma} + \sum_{n} \bs{z}_{fn})
\\
\bs{h}_{n} | \bs{C}_n &\sim \text{Dirichlet}(\bs{\eta} + \sum_{f} \bs{c}_{fn}).
\label{eq:conditionals_DirDir}
\end{align}
As done with the \texttt{Beta-Dir} model, we may use the law of total expectation and the samples $\bs{z}_{fn}^{(j)}$ to obtain Monte-Carlo estimates of the posterior expectations:
\begin{align}
&\mathbb{E}_{\bs{w}_f}[\bs{w}_f | \bs{V}] = 
\mathbb{E}_{\bs{Z}}[\mathbb{E}_{\bs{w}_f}[\bs{w}_f | \bs{V}, \bs{Z}]]
=
\mathbb{E}_{\underline{\bs{Z}}_{f}}[\mathbb{E}_{\bs{w}_f}[\bs{w}_f |  \underline{\bs{Z}}_{f}]]
\approx \frac{\bs{\gamma} + \frac{1}{J} \sum_j \sum_n \mathbf{z}_{fn}^{(j)}}
{\sum_k \gamma_k + N}
\\
&\mathbb{E}_{\bs{h}_{n}}[\bs{h}_{n} | \bs{V}] = 
\mathbb{E}_{\bs{C}}[\mathbb{E}_{\bs{h}_{n}}[\bs{h}_{n}| \bs{V}, \bs{C}]]
=
\mathbb{E}_{\bs{C}_{n}}[\mathbb{E}_{\bs{h}_{n}}[\bs{h}_{n}| \bs{C}_{n}]]
\approx 
\frac{\bs{\eta} + \frac{1}{J} \sum_j\sum_{f} \bs{c}_{fn}^{(j)}}{\sum_k \eta_k +
F}.
\end{align}

As in the \texttt{Beta-Dir} model, $\bs{W}$ and $\bs{H}$ can be sampled in a second step given a collection of samples of $\bs{Z}$ and $\bs{C}$. The predictive posterior and its expectation can be computed as in Eqs. \eqref{eq:predictive_posterior}, \eqref{eq:predictive_expectation}.

\section{Experiments}
We show the performance of the proposed NBMF methods for different tasks in multiple datasets. The datasets, the algorithms, and scripts to replicate all the reported results are available through our R package.

\subsection{Datasets}

\begin{figure}[t]
    \centering
    \begin{tabular}[b]{cc}
        \begin{tabular}[b]{c}
            \begin{subfigure}[t]{0.22\textwidth}
                \includegraphics[scale=0.5]
{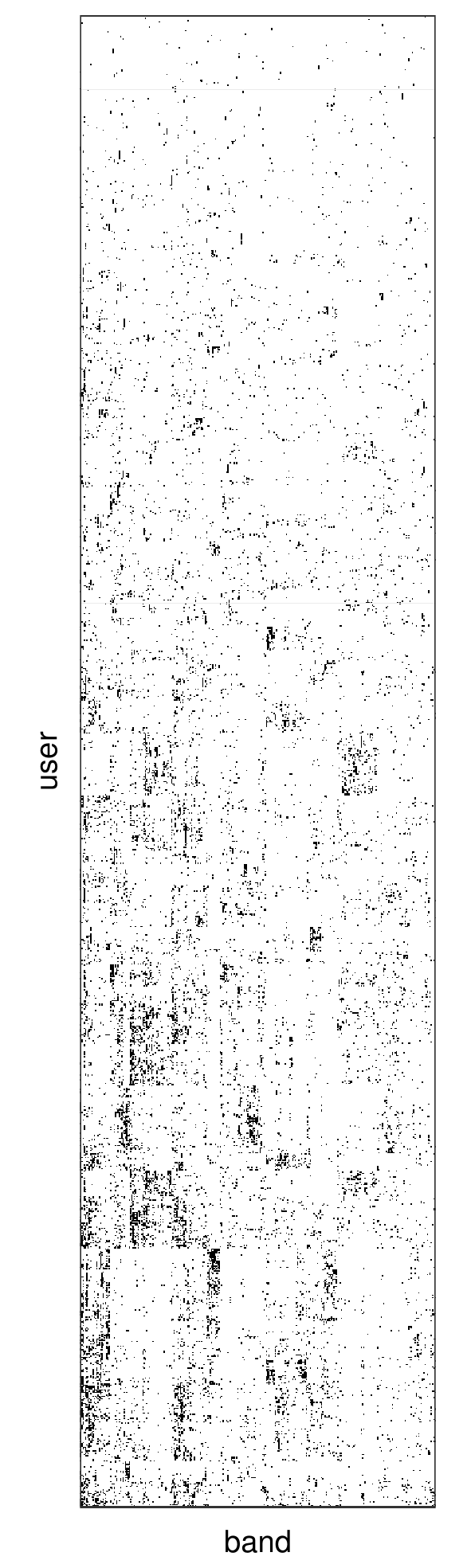} 
                \caption{lastfm} 
            \end{subfigure} 
        \end{tabular}
        &
        \begin{tabular}[b]{c}
            \begin{subfigure}[t]{0.3\textwidth}
                \centering
                \includegraphics[scale=0.55]{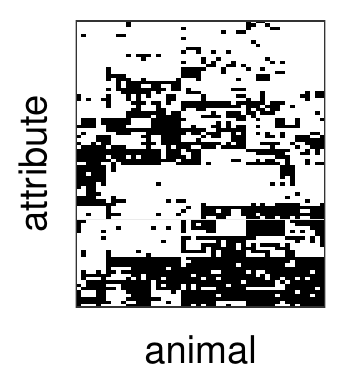}
                \caption{animals}
            \end{subfigure} 
            \begin{subfigure}[t]{0.2\textwidth}
                \centering
                \includegraphics[scale=0.55]{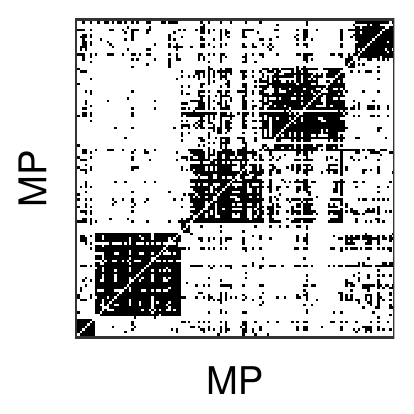}
                \caption{parliament} 
            \end{subfigure}
            \vspace{0.5cm} 
            \\
            \begin{subfigure}[t]{0.65\textwidth}
                \centering
                \includegraphics[scale=0.55]{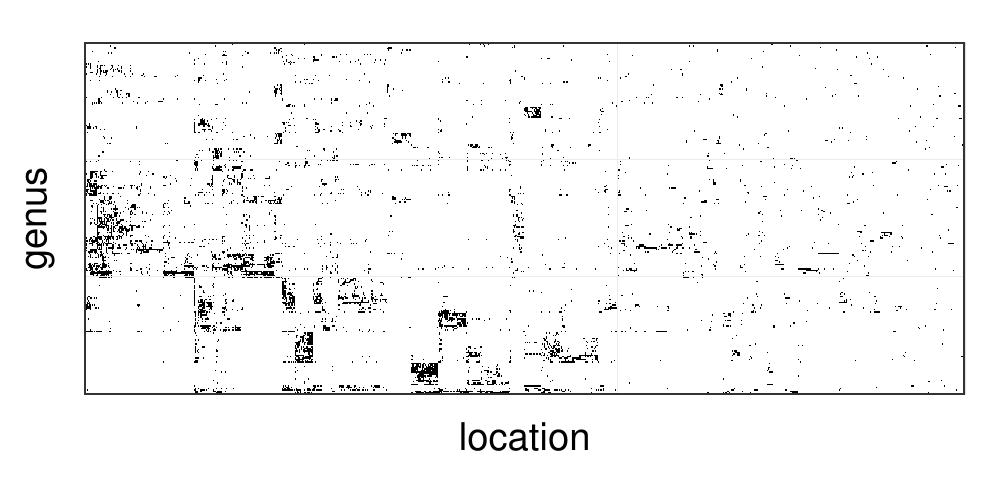}
                \caption{paleo} 
            \end{subfigure}
            \vspace{0.5cm} 
            \\
            \begin{subfigure}[t]{0.65\textwidth}
                \centering
                \includegraphics[scale=0.55]{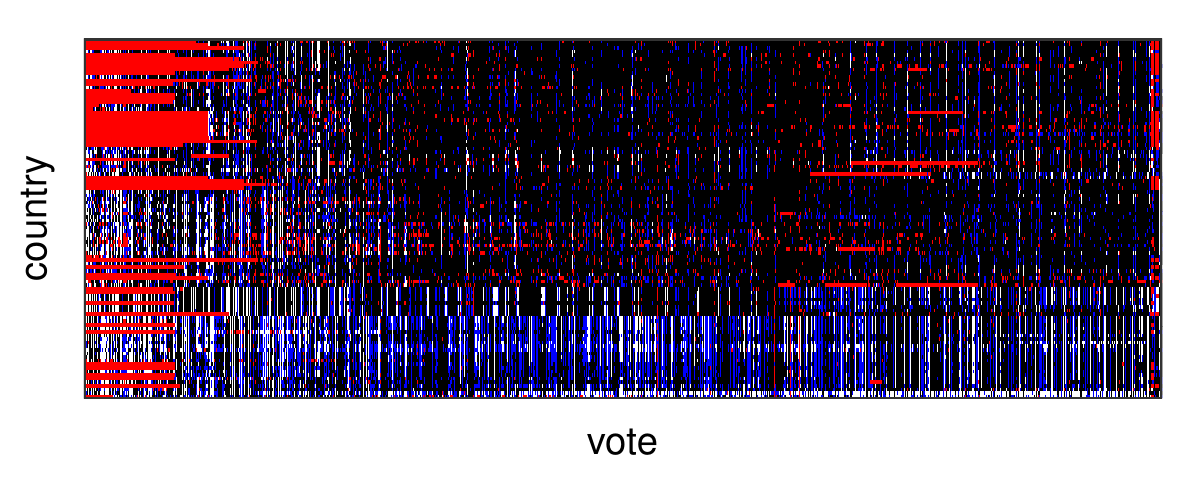}
                \caption{unvote} 
            \end{subfigure}
        \end{tabular}
    \end{tabular}
    \caption{Datasets. Black entries correspond to $v_{fn}=1$ and white entries correspond to zero values. In the {\tt unvote} dataset, blue entries represent ``abstention'' values and red entries are missing votes. For best visualization, rows and columns are re-ordered with complete linkage clustering, except for 1) {\tt parliament} in which parliament members are sorted by parliamentary group and 2) {\tt unvotes} where votes are sorted chronologically.}
\label{fig:originals}
\end{figure}

We consider five different public datasets, described next and displayed in Fig.~\ref{fig:originals}.\\

\noindent \textbf{Animals} (\texttt{animals}). The animals dataset \citep{Kemp2006} contains 50 animals and 85 binary attributes such as \textit{nocturnal}, \textit{hibernates}, \textit{small} or \textit{fast}. The matrix takes $v_{fn}=1$ if animal $n$ has attribute $f$.\\

\noindent \textbf{Last.fm} (\texttt{lastfm}). We use a binarized subset of the Last.fm dataset \citep{Celma2010} where rows correspond to users and columns correspond to musical artists. The matrix takes $v_{fn}=1$ if user $n$ has listened to artist $f$ at least once. The matrix has $F= 285$ rows and $N =1226$ columns.\\

\noindent \textbf{Paleontological data} (\texttt{paleo}). The NOW (New and Old Worlds) fossil mammal database contains information of fossils found in specific paleontological sites \citep{paleo}. From the original paleontological data, we build a matrix where each row is a genus, each column is a location, and $v_{fn}=1$ if genus $f$ has been found at location $n$. We used the same pre-processing as in \cite{Bingham2009} (i.e., we discarded small and infrequent genus, locations with only one genus and kept locations with longitude between 0 and 60 degrees East) and obtained a matrix with $F=253$ rows and $N=902$ columns. \\

\noindent \textbf{Catalan parliament} (\texttt{parliament}). We created a list of the current members of the Catalan parliament and collected the information of who follows whom on Twitter (March 2018). With this data, we created a square adjacency matrix where $v_{fn}=1$ if member $f$ follows member $n$. There are seven political groups represented. The resulting matrix has 135 rows and columns ($F=N$). \\

\noindent \textbf{UN votes} (\texttt{unvotes}). The United Nations General Assembly Voting Data is a dataset that contains the roll-call votes in the UN General Assembly between 1946-2017  \citep{Voeten2012}. Votes can be \textit{yes}, \textit{no}, or \textit{abstention}. From this data, we created a matrix where $v_{fn}=1$ if country $f$ voted \textit{yes} to the call $n$, $v_{fn}=0$ if it voted \textit{no}, and a missing value if the country did not participate in that call or was not a member of the UN at that time. 
Next, abstention votes will be treated as either negative votes (\textit{no}) or missing data, as specified in each experiment. The resulting matrix has $F=200$ rows and $N=5429$ columns.
\subsection{Methods and setting \label{sec:methods}}
\noindent{\bf State-of-the-art methods.} We compare our proposed methods with the following state-of-the-art methods for binary data.

\textbf{\texttt{logPCA-K}}. Probabilistic PCA with Bernoulli likelihood. We use the algorithm presented in \cite{Collins2001}. The notation \texttt{logPCA-K} will embed the chosen number of components $K$ (e.g., \texttt{logPCA-8} signifies $K=8$). We used the R package \texttt{logisticPCA} \citep{Landgraf2015} with default parameters.

\textbf{\texttt{bICA-K}}. The binary ICA method introduced in \cite{Kaban2008}, which uses uncollapsed mean-field variational inference over the partially augmented model (Eqs. \eqref{eq:betadir-aug1}-\eqref{eq:betadir-aug4}). This is also a parametric method that requires setting $K$.\\

\noindent{\bf Proposed methods.} Our proposed methods are as follows.

\textbf{\texttt{Beta-Dir GS}}. Estimation in the \texttt{Beta-Dir} model with collapsed Gibbs sampling. Beta parameters are set to $\alpha_k=\beta_k =1$. To emulate a nonparametric setting, the Dirichlet parameters are set to $\gamma_k = 1/K$ and the number of components is set to $K=100$.

\textbf{\texttt{Beta-Dir VB}}. Estimation in the \texttt{Beta-Dir} model with collapsed variational Bayes (CVB0). Beta parameters are set to $\alpha_k=\beta_k =1$. To emulate a nonparametric setting, we set $\gamma_k = 1/K$ and $K=100$.

\textbf{\texttt{Dir-Dir GS}}. Estimation in the \texttt{Dir-Dir} model with collapsed Gibbs sampling. To emulate a nonparametric setting, we set $\gamma_k=1/K$, $\eta_k=1$, and $K=100$.

\textbf{\texttt{c-bICA-K}}. Collapsed bICA. The algorithm corresponds to $\texttt{Beta-Dir VB}$ with $\alpha_k=\beta_k=\gamma_k=1$. It is the collapsed version of {\texttt{bICA-K}} using CVB0 and \textit{without} the nonparametric approximation.\\

\noindent {\bf Implementation details.} For each dataset, we ran some preliminary experiments to assess the number of iterations needed by the algorithms to converge. For the Gibbs samplers, we set a conservative burn-in phase of $4{,}000$ iterations and kept the last $1{,}000$ samples of $\bs{Z}$ after burn-in. A total number of $500$ iterations where used for the variational algorithms.
In every experiment, we initialized the Gibbs samplers with a random tensor $\bs{Z}$ such that $\bs{z}_{fn} = \ve{e}_{k}$ with random $k$. We did not find a special sensitivity to the initial state of the Gibbs Sampler, but we chose a random initialization as a good practice. We also tried initializing with some variational steps, which is another common practice, but did not see significant improvements. Similarly, we initialized the variational algorithms with a random $\mathbb{E}[\bs{Z}]$ such that $\mathbb{E}[\bs{z}_{fn}] = \ve{e}_{k}$ with random $k$. Our variational algorithms are sensitive to the initial state when some components are empty. Because an empty component has a lower probability of being chosen, in practice the variational algorithms are not capable of refilling it again. Initializing from a random state with lots of used components is, therefore, a safer way to avoid these local maxima.\\

\noindent {\bf Estimators.} The algorithms \texttt{Beta-Dir GS} and \texttt{Dir-Dir GS} return samples from the posterior of $p(\bs{W},\bs{H}|\ve{V})$. Point estimates of the dictionary $\bs{W}$ and data expectation $\hat{\ve{V}} = \bs{W} \bs{H} $ are computed by averaging (posterior mean) and by Eq.~\eqref{eq:predictive_expectation}, respectively. \texttt{Beta-Dir VB}, \texttt{bICA} and \texttt{c-bICA} return variational approximations of the posterior of $\bs{W}$ and $\bs{H}$. Point estimates of $\bs{W}$ and $\hat{\ve{V}}$ are computed from the variational distribution mean and by Eq.~\eqref{eq:predictive_expectation_vb}. \texttt{logPCA} returns maximum likelihood (ML) estimates $\hat{\bs{W}}$ and $\hat{\bs{H}}$. The data expectation is computed as $\hat{\ve{V}} = \sigma(\hat{\ve{W}} \hat{ \ve{H}})$.\\

\noindent {\bf Sensitivity to hyperparameters.} 
When using Dirichlet priors, inferences may be quite sensitive especially for small values of its concentration parameter \citep{Steck2002}. For the setting described in Section \ref{sec:pred}, we have tested our algorithms under toy data generated from the model, setting the concentration parameter to 1. We have repeated different inferences with the concentration parameter of the estimator ranging from 0.1 to 10 and observed very small variations on the perplexity (around 0.1, and decreasing as the size of the observed data increases). Thus, for the sake of simplicity, we have therefore decided to set the concentrations parameters to 1 which corresponds to a uniform Dirichlet prior.\\

\noindent {\bf Computational cost.} 
The time complexity of the collapsed algorithms is $\mathcal{O}(FKN)$ (assuming, for the sake of simplicity, that the Multinomial random number generator is $\mathcal{O}(1)$). Note that, unlike some non-mean-parameterized Bernoulli models \citep{Zhou15}, or all Poisson models, zeros cannot be ignored because they represent another category, not a lack of observation or a zero-count.

\subsection{Dictionary learning and data approximation}

\subsubsection{Experiments with the \texttt{parliament} dataset}

\begin{figure}[t]
\centering
\subcaptionbox{Original}{
\vspace{-0.2cm}
\includegraphics[width=0.21\textwidth]{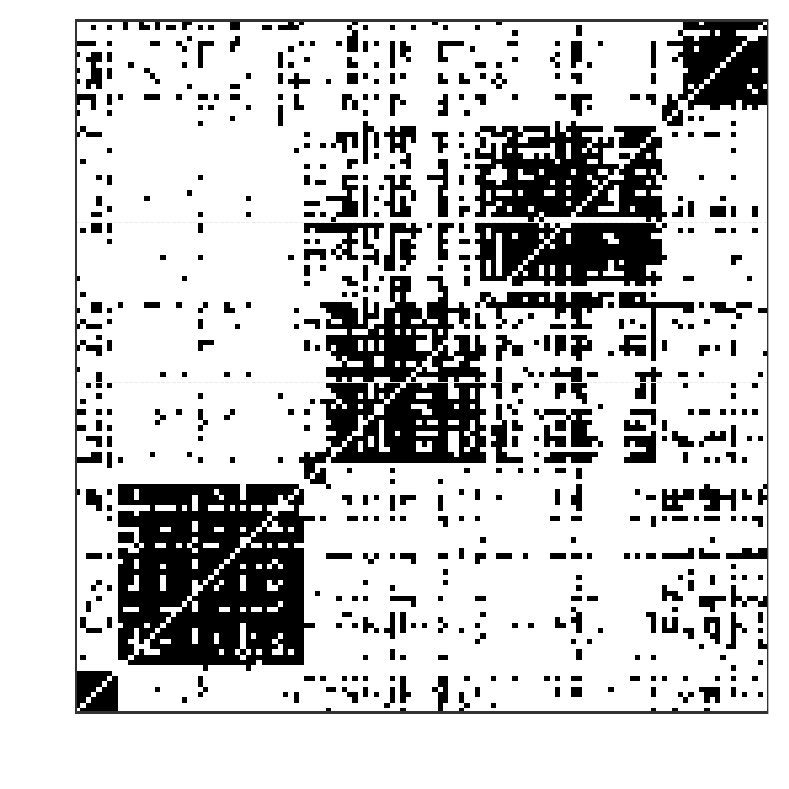}}\\
\vspace{0.2cm}
\subcaptionbox{Beta-Dir GS}{\includegraphics[width=0.19\textwidth]
{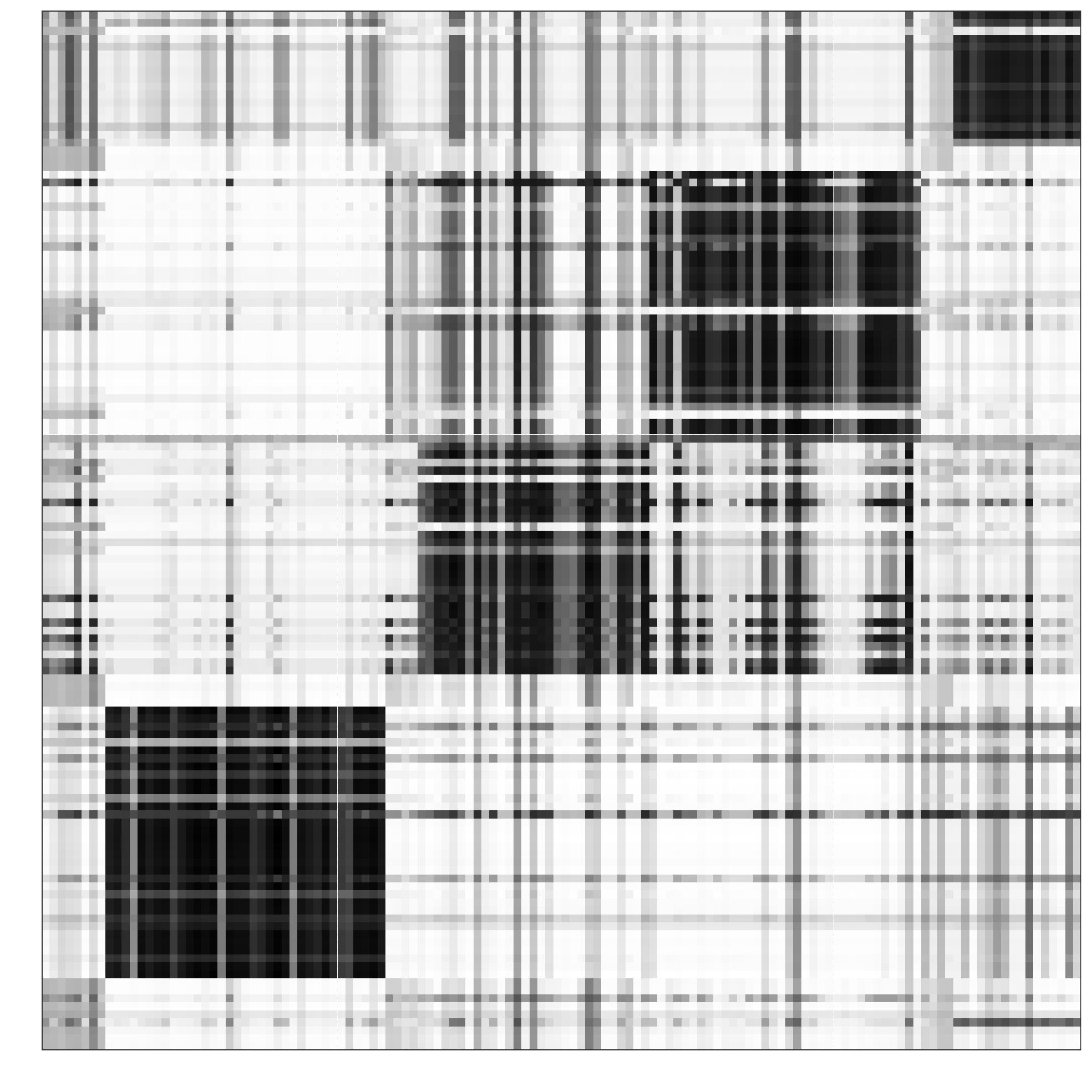}}
\subcaptionbox{Beta-Dir VB}{\includegraphics[width=0.19\textwidth]{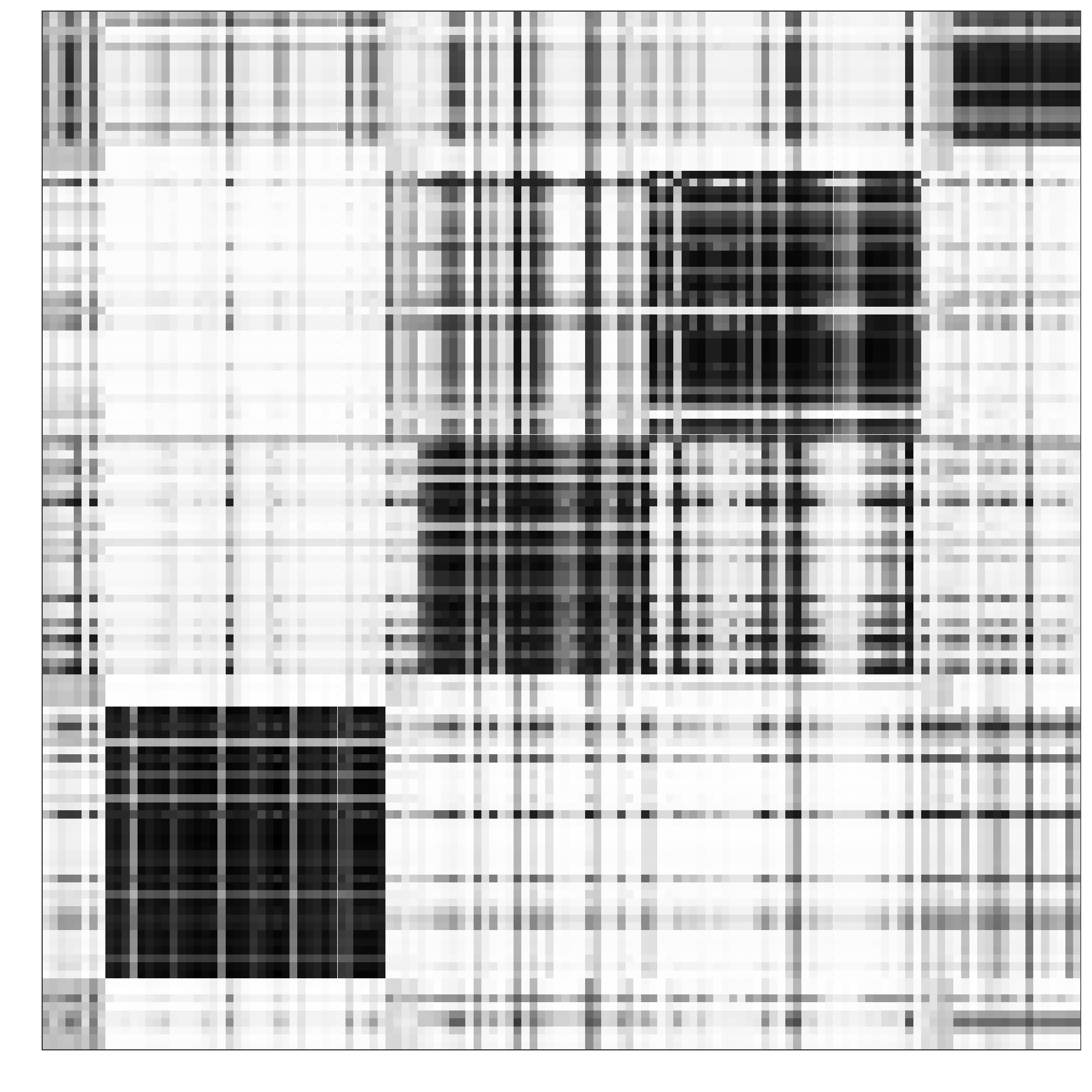}}
\subcaptionbox{Dir-Dir GS}{\includegraphics[width=0.19\textwidth]{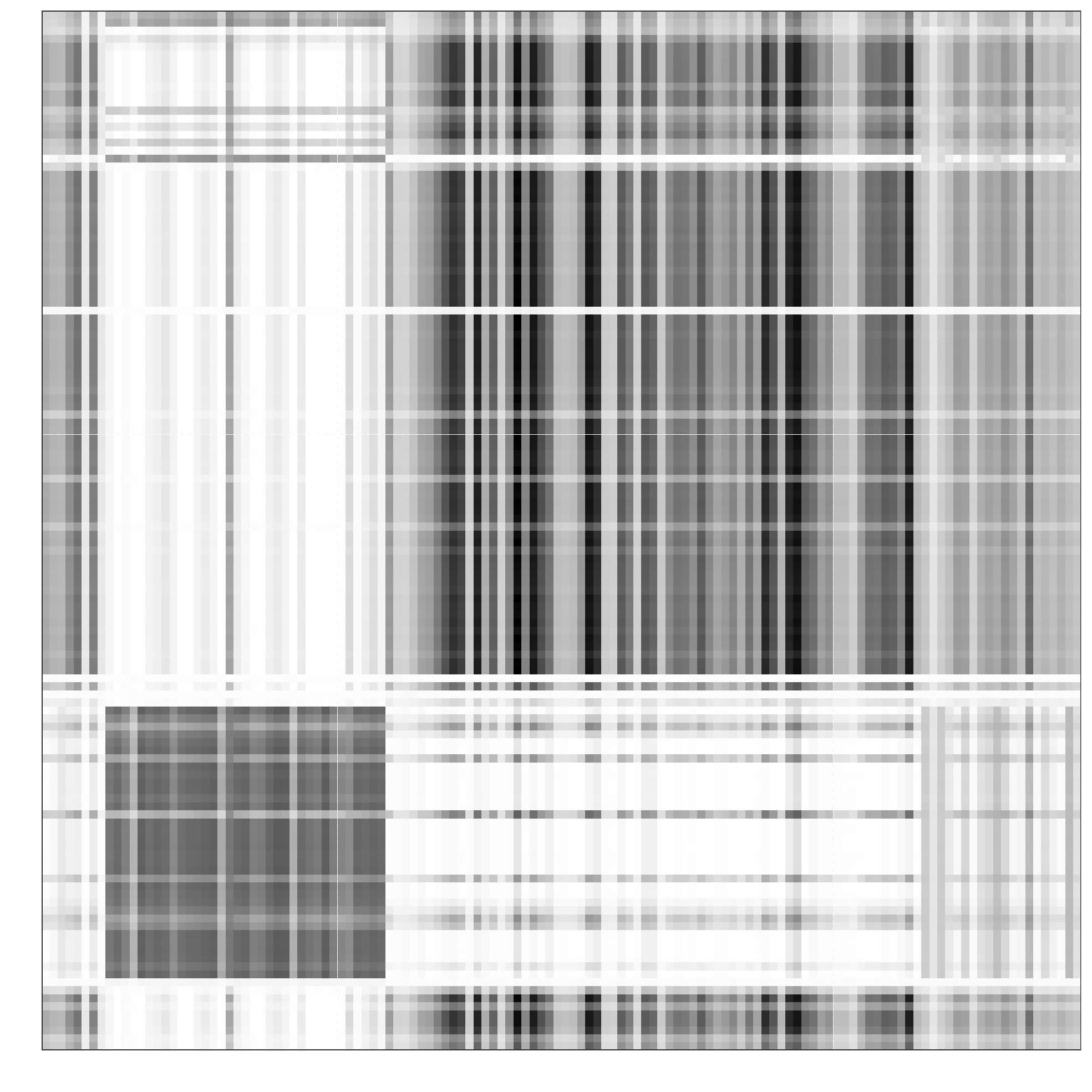}}
\subcaptionbox{bICA-8}{\includegraphics[width=0.19\textwidth]{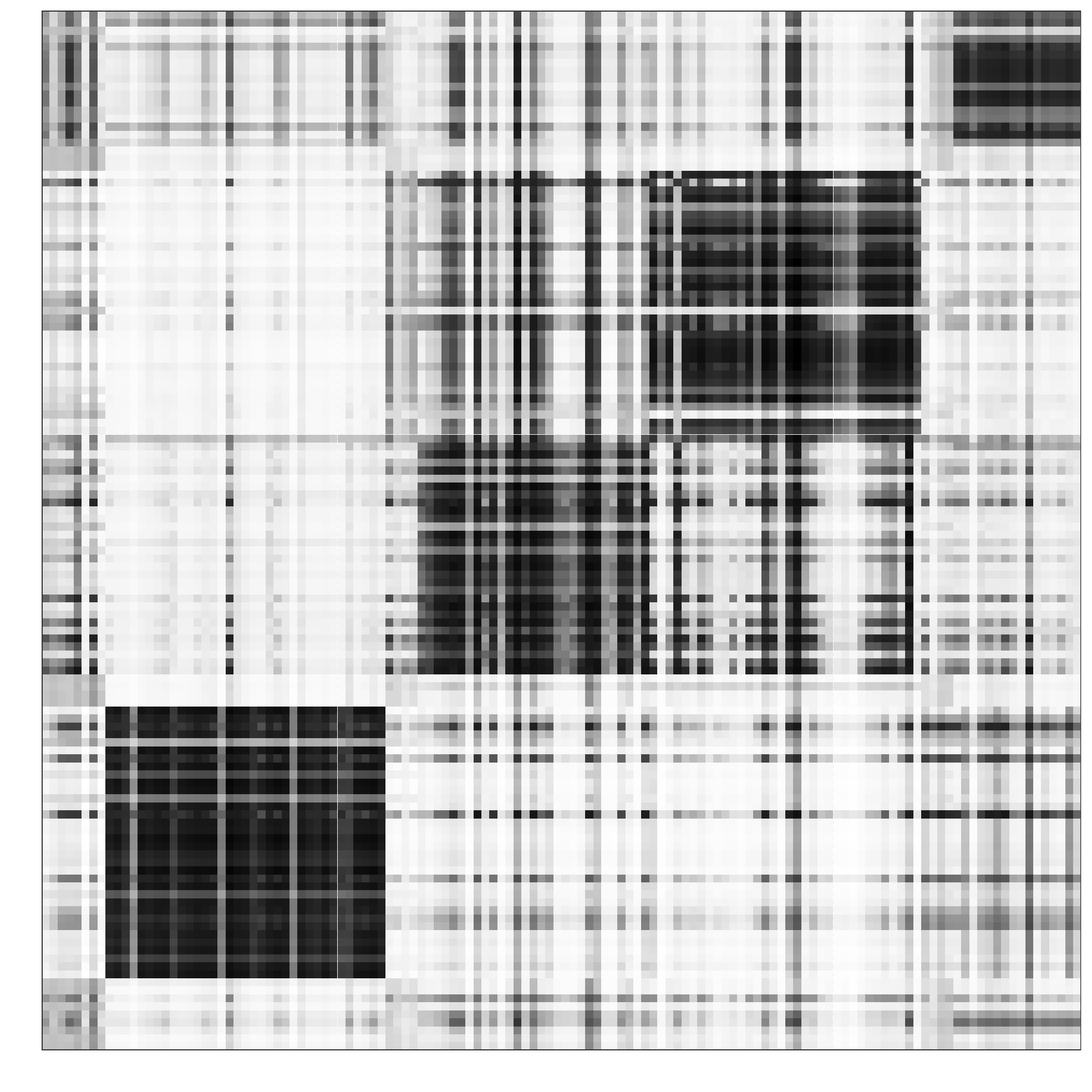}}
\subcaptionbox{logPCA-8}{\includegraphics[width=0.19\textwidth]{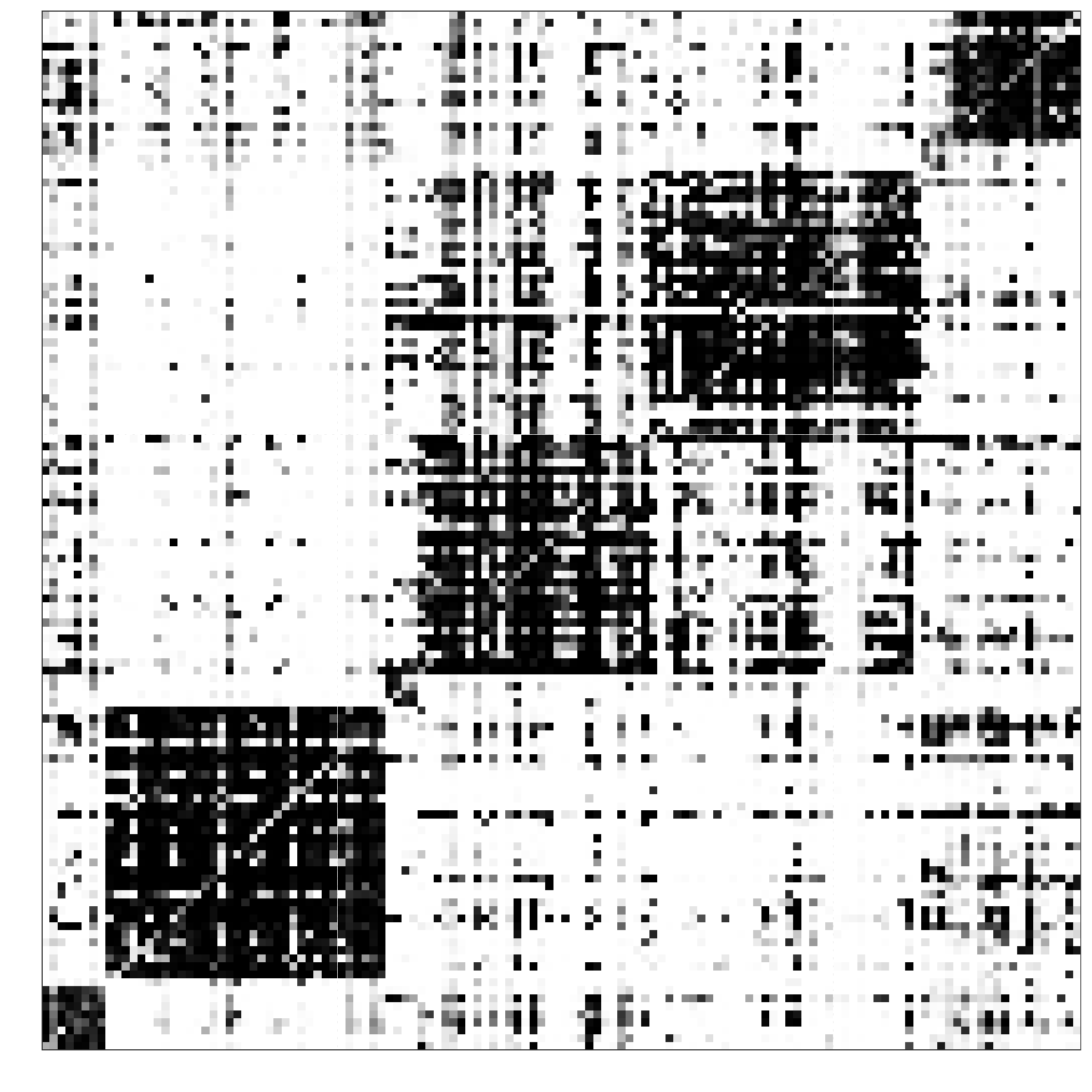}}
\caption{Reconstructed matrices for the \texttt{parliament} dataset.}
\label{fig:reconstructions}
\end{figure}

\begin{figure*}
\centering
\includegraphics[width=0.75\textwidth]{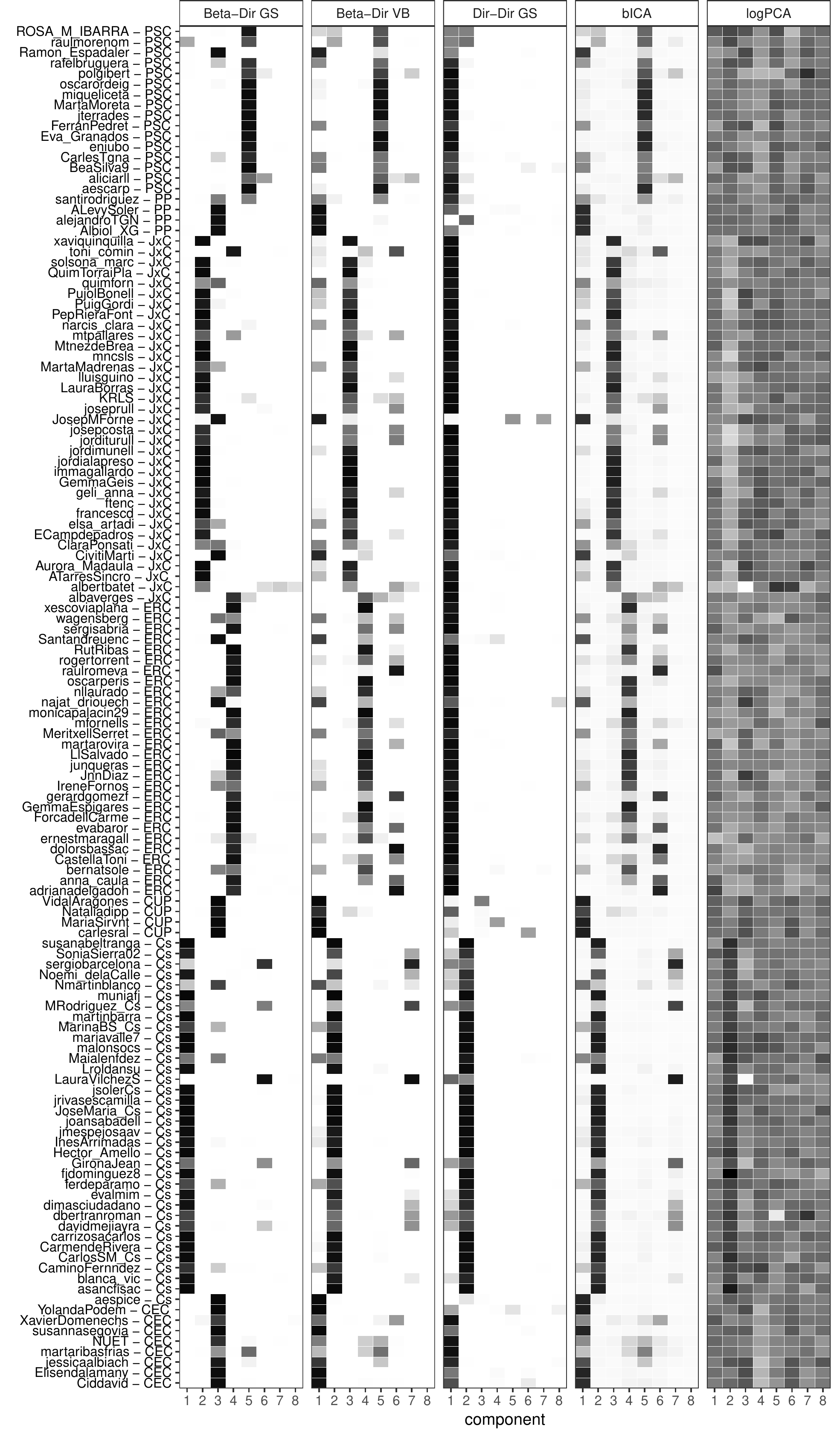}
\caption{Estimated dictionaries from the \texttt{parliament} dataset. Members are sorted by party and then alphabetically. Columns are sorted by their norm. Only the first eight columns are displayed for the nonparametric methods \texttt{Beta-Dir GS}, \texttt{Beta-Dir VB} and \texttt{Dir-Dir GS}. The results displayed for \texttt{bICA} and \texttt{logPCA} are with $K=8$. The values of $\bs{W}$ estimated by \texttt{logPCA-8} belong to the $[-207.17, 217.92]$ range and have been linearly mapped to the $[0,1]$ range for visual display.}
\label{fig:dictionary_parliament}
\end{figure*}

First, we want to form an idea of how well the different models can fit original data. We focus on the \texttt{parliament} dataset, which has reasonable size and a clear structure. We applied the three proposed nonparametric methods \texttt{Beta-Dir GS}, \texttt{Beta-Dir VB}, \texttt{Dir-Dir GS} and the state-of-the-art methods \texttt{bICA} and \texttt{logPCA}.  For each method, we compute the negative log-likelihood of the data approximation $\hat{\ve{V}}$, which serves as a measure of fit:
\begin{align}
D(\ve{V} | \hat{\ve{V}}) = - \sum_{fn} \log p(v_{fn} | \hat{v}_{fn}).
\end{align}
\texttt{bICA} was run with increasing values of $K$ and the fit ceased increasing for $K=8$ which is the value used in the results (note that in this case $\hat{\ve{V}}$ is the posterior mean estimate and not the ML estimate, so the likelihood is not meant to increase monotonically). $\texttt{logPCA}$ was run with the same value $K=8$. {The data approximations $\hat{\ve{V}}$ and dictionaries obtained with the different methods are displayed in Figs.~\ref{fig:reconstructions} and~\ref{fig:dictionary_parliament}, respectively.}\\

In terms of data approximation, \texttt{Beta-Dir VB} achieves the best fit among the mean-parameterized models in terms of negative log-likelihood ($4{,}729$) followed by \texttt{Beta-Dir GS} ($4{,}863$). The dictionaries returned by these two algorithms are very similar, with only nine active components. \texttt{bICA-8} comes next in terms of fit ($4{,}957$). We also applied \texttt{bICA} with $K>8$ components but this did not substantially improve the likelihood. \texttt{Dir-Dir GS} returns the worst fit ($8{,}930$), with only two active components. Overall \texttt{logPCA-8} returns the smallest negative log-likelihood ($1{,}783$). This due to its larger flexibility as compared to the mean-parameterized models (real-valued factors $\bs{W}$ and $\bs{H}$, with product $\bs{W} \bs{H}$ mapped to $[0,1]$). However, this is at the cost of meaningfulness of the decomposition, as shown in Fig.~\ref{fig:dictionary_parliament} {and explained next.}\\

In dictionary learning, we want to learn a meaningful decomposition of the data. The columns of the dictionary $\bs{W}$ are expected to contain {\em patterns} or {\em prototypes} characteristic of the data. In particular, NMF is known to produce so-called part-based representations (each sample, a column of $\bs{V}$, is approximated as a constructive linear combination of building units) \citep{lee99}. When the rows of the dictionary are given Dirichlet priors, $\ve{w}_{f}$ can also be interpreted as the probability distribution of feature $f$ over the $K$ components. In Fig.~\ref{fig:dictionary_parliament}, the rows of the dictionaries displayed correspond to members of the parliament (MP). For each MP we show its Twitter username and its political party. The dictionaries returned by the mean-parameterized factorization methods are easily interpretable. In particular, the dictionaries returned by \texttt{Beta-Dir GS}, \texttt{Beta-Dir VB} and to some extent \texttt{bICA-8} closely reflect the party memberships of the MPs. \texttt{Dir-Dir GS}, which is based on a less flexible model, only captures two sets of MPs, one with the members of {\em Cs} (the main opposition party) and the other with members of the remaining parties, regardless of political alignment (left-wing, right-wing, independentist and anti-independentist). In contrast, the dictionary returned by \texttt{logPCA-8} is much more difficult to interpret.

\subsubsection{Experiments with the \texttt{unvotes} dataset}

\begin{figure*}
\centering
\includegraphics[width=0.75\textwidth]{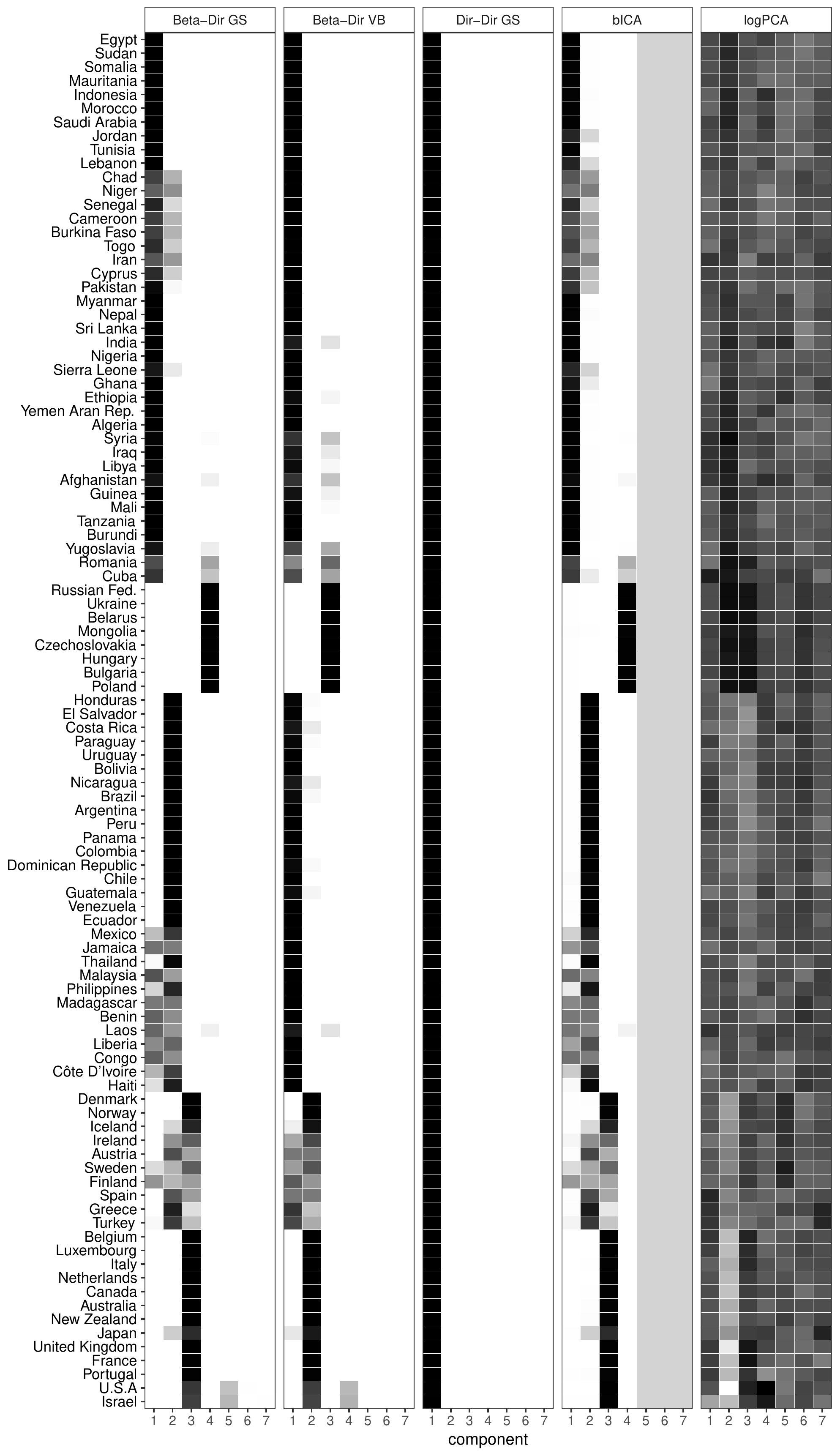}
\caption{Estimated dictionaries from the \texttt{unvotes} dataset. Columns are sorted by their norm. Only the first seven columns are displayed for the nonparametric methods \texttt{Beta-Dir GS}, \texttt{Beta-Dir VB} and \texttt{Dir-Dir GS}. The results displayed for \texttt{bICA} and \texttt{logPCA} are with $K=4$ and $K=7$, respectively. The values of $\bs{W}$ estimated by \texttt{logPCA-7} belong to the $[-629.7, 335.1]$ range and have been linearly mapped to the $[0,1]$ range for visual display.}
\label{fig:dictionary_un}
\end{figure*}

In this section, we consider a subset of \texttt{unvotes}, reduced to the 1946-1990 range which corresponds to the Cold War period. Furthermore, the abstentions are here treated as missing values. Fig.~\ref{fig:dictionary_un} shows the dictionaries learned by the five considered methods. As before, \texttt{bICA} was applied with various values of $K$ and we selected the value that leads to the smallest negative log-likelihood ($K=7$). Accordingly, \texttt{logPCA} was also applied with $K=7$. 

Fig.~\ref{fig:dictionary_un} shows that \texttt{Beta-Dir GS} returns the finest dictionary, detecting political blocks that tended to vote similarly in the UN assembly and capturing some nuances that the other algorithms do not find. European countries (and members or allies of NATO such as the USA, Japan, or Australia) are concentrated in one component, denoting similar voting strategies. The former members of the Soviet Union and the Warsaw Pact also form a block of their own, with some allies such as Cuba or the former Yugoslavia. Members of the Non-Aligned Movement (even countries that became members after 1991, such as Guatemala, Thailand, or Haiti), from Egypt to Cuba and from Honduras to Haiti, are split into two blocks: the Latin American group and the Asian-African group. Another detected alliance is between the United States and Israel, which are distributed between the European component and a component of their own. \texttt{Beta-Dir VB} detects the split between the Warsaw and NATO blocks, and the alliance between the USA and Israel, but it fails to detect the two subgroups of the Non-Aligned Movement, which is considered a single block. \texttt{bICA-7} returns similar results to \texttt{Beta-Dir GS} but fails to detect the alliance between the USA and Israel. Note that the results of \texttt{bICA} are obtained with a well-chosen value of $K$ while \texttt{Beta-Dir GS} automatically detects a suitable value. The underlying assumption of \texttt{Dir-Dir} (one topic per country and one topic per vote) seems too simplistic for this dataset, and the algorithm puts every country in the same component. Again and as somewhat expected, the dictionary learned by \texttt{logPCA} is more difficult to interpret.

\subsubsection{Experiments with the \texttt{paleo} dataset}

\begin{figure*}
\centering
\includegraphics[width=0.75\textwidth]
{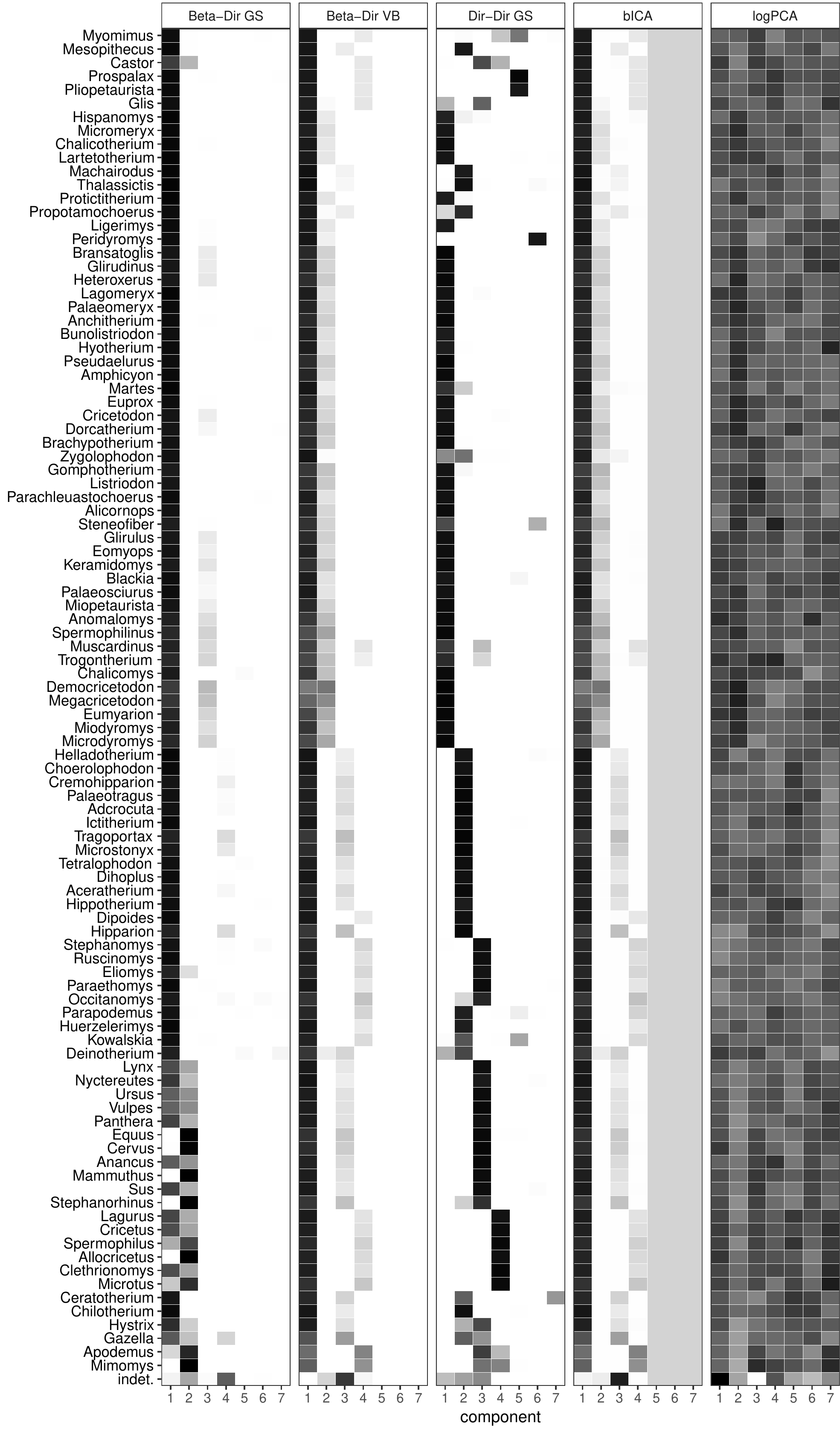}
\caption{
Estimated dictionaries from the \texttt{paleo} dataset. Columns are sorted by their norm. Only the first seven columns are displayed for the nonparametric methods \texttt{Beta-Dir GS}, \texttt{Beta-Dir VB} and \texttt{Dir-Dir GS}. The results displayed for \texttt{bICA} and \texttt{logPCA} are with $K=4$ and $K=7$, respectively. The values of $\bs{W}$ estimated by \texttt{logPCA-7} belong to the $[-102.3, 118.7]$ range and have been linearly mapped to the $[0,1]$ range for visual display.}
\label{fig:dictionary_paleo}
\end{figure*}

We finally look into the dictionaries returned by the five considered method on the \texttt{paleo} dataset, see Fig.~\ref{fig:dictionary_paleo}. The same strategy was applied to find a suitable value of $K$ for \texttt{bICA} and \texttt{logPCA}, leading to $K=7$. The results can be read as the probability of a genus to be found in a set of prototypical locations. Interestingly, \texttt{Dir-Dir GS} is the method that returns the most detailed dictionary for this dataset. The other methods tend to produce larger clusters of genera. This highlights the importance of choosing the right model for each dataset since they imply different underlying assumptions. \texttt{Dir-Dir GS} assumes one topic per genus and one topic per location. We will see in the following section that \texttt{Dir-Dir GS} also gives the best predictions for this dataset. Again, the dictionary obtained with \texttt{logPCA} is harder to interpret.

\subsection{Prediction \label{sec:pred}}

\subsubsection{Experimental setting}

We now evaluate the capability of the five previously considered methods together with \texttt{c-bICA} to predict missing data. For each of the five considered datasets, we applied the algorithms to a $75\%$ random subset of the original data. We here use the full \texttt{unvotes} in which abstentions are treated as negative votes ($v_{fn}=0$). \texttt{bICA} and \texttt{c-bICA} where applied with $K=2,\ldots, 8$. \texttt{logPCA} was applied with $K=2,\ldots, 4$. Then we computed the {\em perplexity} of the test set (the 25\% held-out entries) given the estimate $\hat{\bs{V}} = \mathbf{E}[\bs{WH} | \bs{V}_\text{train}]$ (for all methods except \texttt{logPCA-K}) or $\hat{\bs{V}} = \sigma(\hat{\bs{W}} \hat{ \bs{H}})$ (for \texttt{logPCA}). The perplexity is here simply taken as the negative log-likelihood of the test set \citep{Hofmann1999a}:
\begin{align}
perplexity = - \frac{1}{T} \sum_{(f,n) \in \text{test}} \log p(v_{fn} | \hat{\bs{V}})
\end{align}
where $T$ is the number of elements in the test set (in our case, $T = 0.25 F N$).

\subsubsection{Prediction performance}
\begin{figure}[!tb]
\centering
\includegraphics[width=0.65\textwidth]
{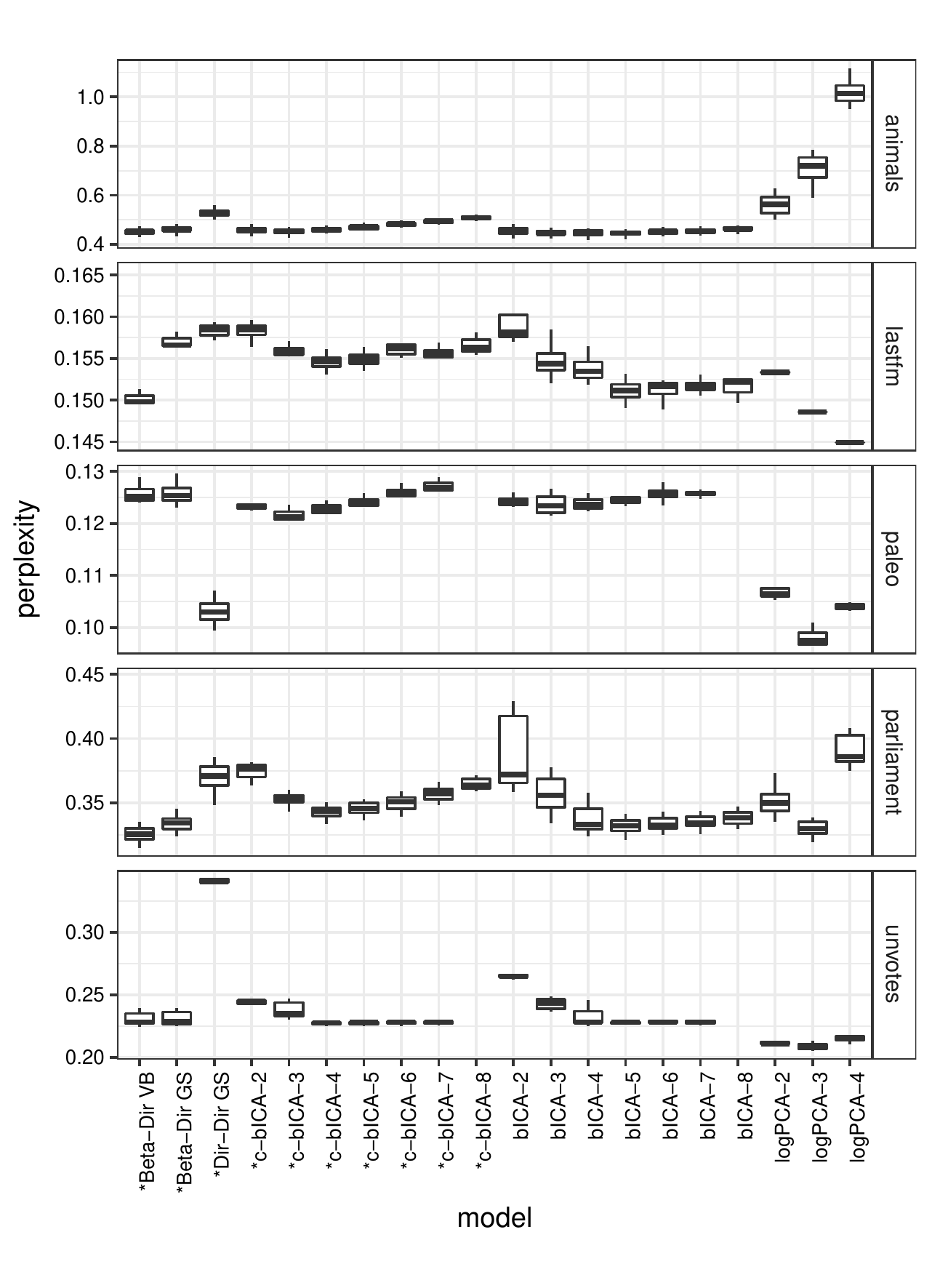}
\caption{Prediction performance measured by perplexity (lower values are better). The methods introduced in this paper are marked with an asterisk.}
\label{fig:perplexity}
\end{figure}

Fig.~\ref{fig:perplexity} displays the perplexities obtained by all methods from 10 repetitions of the experiment with randomly selected training and test sets, and random initializations (the same starting point is used for \texttt{bICA} and \texttt{c-bICA}). The proposed \texttt{Beta-Dir VB} performs similarly or better (\texttt{lastfm}, \texttt{parliament}) than \texttt{bICA}, while automatically adjusting the number of relevant components. As hinted from the dictionary learning experiments, \texttt{Dir-Dir GS} performs considerably better than the other mean-parameterized methods on the \texttt{paleo} dataset. 
{\texttt{c-bICA} does not specifically improve over \texttt{bICA} (remember they are based on the same model, only inference changes) and performs worse in some cases (\texttt{lastfm}, \texttt{parliament}). However, its performance is more stable, with less variation between different runs, a likely consequence of the collapsed inference.} 

Despite its flexibility (unconstrained $\bs{W}$ and $\bs{H}$), \texttt{logPCA} provides marginally better perplexity (except on the {\tt animal} dataset where it performs worse than almost all other methods), and only given a suitable value of $K$. Its predictive performance can drastically decay with ill-chosen values of $K$. In contrast, our proposed methods do not require tuning $K$ to a proper value. Furthermore, they provide competitive prediction performance together with the interpretability of the decomposition. We have also compared against a standard \texttt{LDA} using the same range of $K$ than \texttt{logPCA}, but the perplexity was much worse (around six) for all datasets; we did not plot it due to the high difference in the scale.

\subsubsection{Convergence of the variational inference algorithms}

\begin{figure}[!t]
\centering
\includegraphics[width=0.65\textwidth]{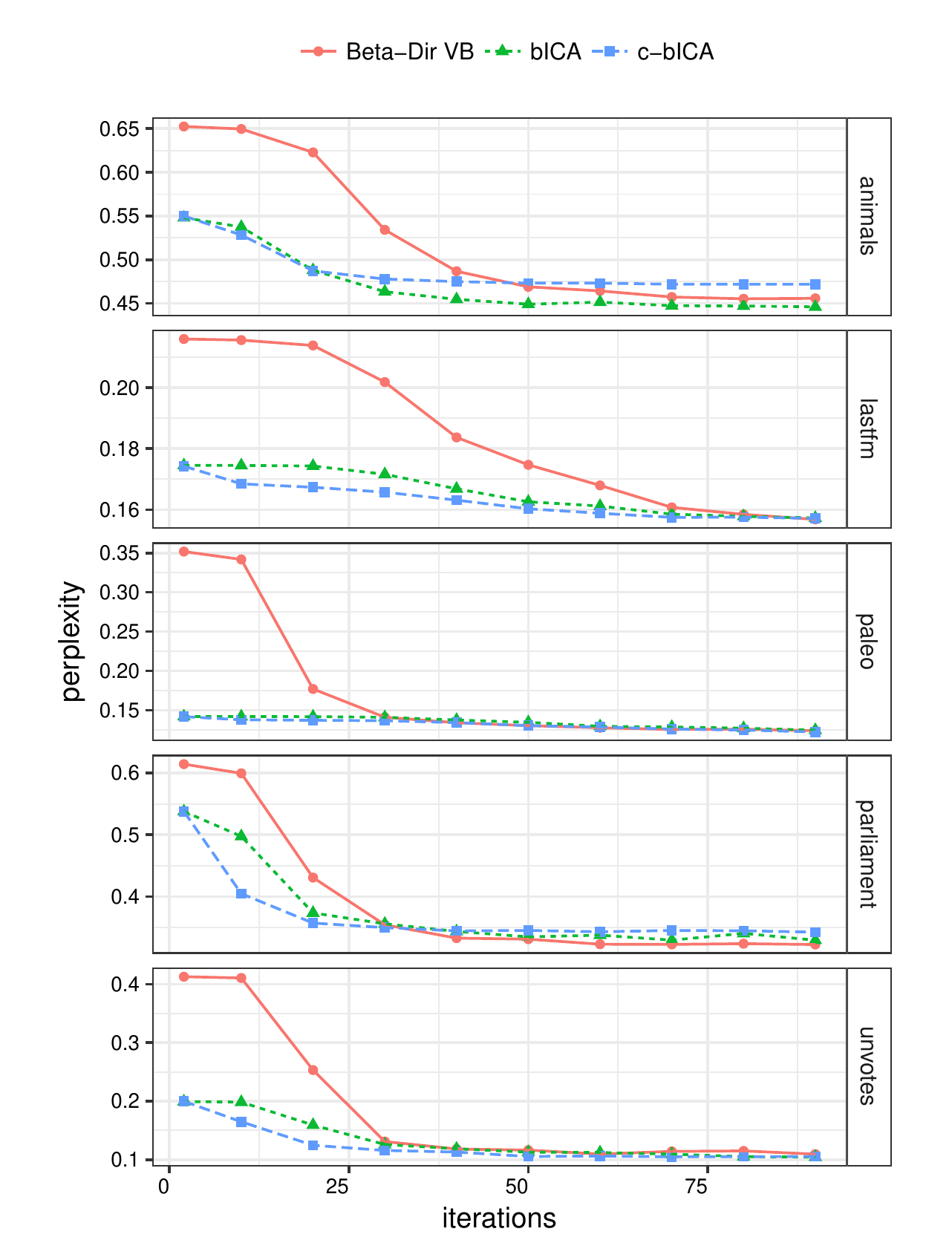}
\caption{Average perplexity (over 10 repetitions) returned by the variational inference algorithms along iterations.}
\label{fig:convergence}
\end{figure}

Fig.~\ref{fig:convergence} displays the average perplexity values returned by the variational algorithms \texttt{Beta-Dir VB}, \texttt{c-bICA-5} and \texttt{bICA-5} along iterations. As expected, \texttt{c-bICA} tends to converge faster than \texttt{bICA} though not consistently so. Being initialized with a full tensor of dimension $K=100$ (as described in Section~\ref{sec:methods}), \texttt{Beta-Dir VB} starts with a relatively higher perplexity but catches up with the two other methods in a reasonable number of iterations.

\section{Conclusions}
We have presented a unified view for Bayesian mean-parameterized NBMF. The interest of mean-parameterized models in NMF is that they keep factors interpretable since they belong to the same space than the observed data. We have addressed three models that correspond to three possible sets of constraints that each respect mean-parameterization. One model, \texttt{Dir-Beta}, is a Bayesian extension of the Aspect Bernoulli model of \cite{Bingham2009}. Another model, \texttt{Beta-Dir}, corresponds to the binary ICA model of \cite{Kaban2008}. We have proposed a new collapsed Gibbs sampler and a new collapsed variational inference method for estimation in these models. We have proposed a novel, third model, \texttt{Dir-Dir}, and we have designed a collapsed Gibbs sampler for inference with this model. Lastly, we have proposed a nonparametric extension for these three models. 
Experiments have shown that our nonparametric methods can achieve similar performance than the state-of-the-art methods applied with a suitable value of $K$. As expected, the more flexible \texttt{logPCA} can achieve better data approximation and in some cases prediction, but at the cost of interpretation which of utter importance in some applications.

\section*{Acknowledgements} This work has received funding from the European Research Council (ERC) under the European Union's Horizon 2020 research and innovation programme under grant agreement No 681839 (project FACTORY).

\bibliographystyle{spbasic} 
\bibliography{BernoulliNMF}

\appendix
\section{Probability distributions functions \label{app:dist}}

\subsection{Bernoulli distribution}
Distribution over a binary variable $x \in \{0,1\}$, with mean parameter $\mu \in [0,1]$:
\begin{align}
\text{Bernoulli}(x | \mu) &= \mu^x(1-\mu)^{1-x}.
\end{align}

\subsection{Beta distribution}
Distribution over a continuous variable $x \in [0,1]$, with shape parameters $a >0$, $b>0$:
\begin{align}
\text{Beta}(x | a,b)&= \frac{\Gamma(a+b)}{\Gamma(a)\Gamma(b)}x^{a-1}(1-x)^{b-1}.
\end{align}

\subsection{Gamma distribution}
Distribution for a continuous variable $x >0$, with shape parameter $a >0$ and rate parameter $b>0$:
\begin{align}
\text{Gamma}(x | a,b)&= \frac{b^a}{\Gamma(a)}x^{a-1} e^{-bx}.
\end{align}

\subsection{Dirichlet distribution}
Distribution for $K$ continuous variables $x_{k} \in [0,1]$ such that $\sum_k x_k = 1$. Governed by $K$ shape parameters $\alpha_1,...\alpha_K$ such that $\alpha_k>0$:
\begin{align}
\text{Dirichlet}(\bs{x} | \bs{\alpha}) &=
\frac{\Gamma(\sum_k \alpha_k)}{\prod_k \Gamma(\alpha_k)}\prod_k x^ {\alpha_k-1}.
\end{align}

\subsection{Discrete distribution}
Distribution for the discrete variable $\ve{x} \in \{\ve{e}_{1}, \ldots, \ve{e}_{K} \}$, where $\ve{e}_{i}$ is the $i^{th}$ canonical vector. Governed by the discrete probabilities $\mu_1,...,\mu_K$ such that $\mu_{k} \in [0,1]$ and $\sum_k \mu_k = 1$:
\begin{align}
p(\ve{x} = \ve{e}_{k}) = \mu_{k}
\end{align}
The probability mass function can be written as:
\begin{align}
\text{Discrete}(\bs{x} | \bs{\mu}) & = \prod_k \mu_k^{x_k}.
\end{align}
We may write $\text{Discrete}(\bs{x} | \bs{\mu}) = \text{Multinomial}(\bs{x} | 1, \bs{\mu})$.

\subsection{Multinomial distribution}
Distribution for an integer-valued vector $\bs{x}=[x_1,...,x_K]^T \in \mathbb{N}^K$. Governed by the total number $L = \sum_k x_k$ of events assigned to $K$ bins and the probabilities $\mu_k$ of being assigned to bin $k$:
\begin{align}
\text{Multinomial}(\bs{x} | L, \bs{\mu}) & = \frac{L!}{x1!...x_K!}\prod_k \mu_k^{x_k}.
\end{align}

\section{Derivations for the \texttt{Beta-Dir} model}\label{sec:appendix_BetaDir}

\subsection{\new{Marginalizing out $\bs{W}$ and $\bs{H}$ from the joint likelihood}}\label{sec:appendix_BetaDir_marginals}

\new{
{We seek to compute marginal joint probability introduced in Eq.~\eqref{eq:marginalized_BetaDir} and given by:}
\begin{align*}
p(\mathbf{V}, \mathbf{Z}) =
\prod_f 
\overbrace{
\int
p(\bs{w}_{f})
\prod_{n}
p(\bs{z}_{fn} | \bs{w}_f) \,d\mathbf{w}_f
}^{p(\underline{\bs{Z}}_{f})
}
\prod_n
\overbrace{
\int
\prod_{k} 
p(h_{kn})
\prod_{f}
p(v_{fn} | \bs{h}_n, \bs{z}_{fn})
\, d\ve{h}_{n}
}^{p(\bs{v}_n | \bs{Z}_n)
}.
\label{eq:app_app_marginalized_BetaDir}
\end{align*}
{Using the expression of the normalization constant of the Dirichlet distribution, the first integral can be computed as follows:}
\begin{align}
p(\underline{\bs{Z}}_{f}) &= 
\int
p(\bs{w}_{f})
\prod_{n}
p(\bs{z}_{fn} | \bs{w}_f) \,d\mathbf{w}_{f}
\\&=
\int
\frac{\Gamma(\sum_k \gamma_k)}{\prod_k \Gamma(\gamma_k)}\prod_k w_{fk}^{\gamma_k-1}\prod_n w_{fk}^{z_{fkn}} \,d\mathbf{w}_{f}
\\&=
\frac{\Gamma(\sum_k \gamma_k)}{\prod_k \Gamma(\gamma_k)}
\int
\prod_k w_{fk}^{\gamma_k + L_{fk}-1} \,d\mathbf{w}_{f}
\\&=
\frac{\Gamma(\sum_k \gamma_k)}{\prod_k \Gamma(\gamma_k)}
\frac{\prod_k \Gamma(\gamma_k +  L_{fk})}{\Gamma(\sum_k \gamma_k + L_{fk})}.
 \end{align}
{The second integral in Eq.~\eqref{eq:marginalized_BetaDir} is computed as follows. In Eq.~\eqref{eq:BernBern} we use that $p(v_{fn} | \bs{h}_n, \bs{z}_{fn}) = \text{Bernoulli}(v_{fn} | \prod_k h_{kn}^{z_{fkn}}) = \prod_k \text{Bernoulli}(v_{fn}|h_{kn})^{z_{fkn}}$ {(recall that $\mathbf{z}_{fn}$ is an indicator vector)}. In Eq.~\eqref{eq:vZend}, we use the expression of the normalization constant of the Beta distribution.
}
\begin{align}
p(\bs{v}_n | \bs{Z}_n) &= 
\int
\prod_k
p(h_{kn})
\prod_{f}
p(v_{fn} | \bs{h}_n, \bs{z}_{fn})
\, {d \mathbf{h}_n}
\\&=
\int \prod_k 
\left[
\frac{\Gamma(\alpha_k + \beta_k)}{\Gamma(\alpha_k)\Gamma(\beta_k)}
h_{kn}^{\alpha_k-1}(1-h_{kn})^{\beta_k-1}
\right]
\prod_{fk} \left[h_{kn}^{v_{fn}}(1-h_{kn})^{1-v_{fn}}\right]^{z_{fkn}} {d \mathbf{h}_n} \label{eq:BernBern}
\\&=
\prod_k \int  
\frac{\Gamma(\alpha_k + \beta_k)}{\Gamma(\alpha_k)\Gamma(\beta_k)}
h_{kn}^{\alpha_k-1}(1-h_{kn})^{\beta_k-1}
\prod_{f} \left[h_{kn}^{v_{fn}}(1-h_{kn})^{1-v_{fn}}\right]^{z_{fkn}} dh_{kn}
\\&=
\prod_k
\frac{\Gamma(\alpha_k + \beta_k)}{\Gamma(\alpha_k)\Gamma(\beta_k)}
\int
h_{kn}^{\alpha_k + A_{kn} -1}(1-h_{kn})^{\beta_k + B_{kn}-1}
dh_{kn}
\\&=
\prod_k
\frac{\Gamma(\alpha_k + \beta_k)}{\Gamma(\alpha_k)\Gamma(\beta_k)}
\frac{\Gamma(\alpha_k + A_{kn})\Gamma(\beta_k + B_{kn})}{\Gamma(\alpha_k + \beta_k + M_{kn})}. \label{eq:vZend}
\end{align}
}

\subsection{Conditional prior and posterior distributions of $\bs{z}_{fn}$}\label{sec:appendix_BetaDir_conditionals}
Applying the Bayes rule, the conditional posterior of $\bs{z}_{fn}$ is given by: 
\begin{align}
p(\bs{z}_{fn} | \bs{Z}_{\neg fn}, \bs{V}) \propto p(\bs{V} | \bs{Z})p(\bs{z}_{fn} | \bs{Z}_{\neg fn}). \label{eqn:cond1}
\end{align}
The likelihood itself decomposes as $p(\bs{V} | \bs{Z}) = \prod_{n} p(\bs{v}_n | \bs{Z}_n)$ and we may ignore the terms that do not depend on $\bs{z}_{fn}$. Using Eq.~\eqref{eq:marginalized_BetaDir_V} and the identity $\Gamma(n + b) = \Gamma(n)n^b$ where $b$ is a binary variable, we may write:
\begin{align}
p(\bs{v}_n | \bs{Z}_n)
&=
\prod_{k}
\frac{\Gamma(\alpha_k + \beta_k)}{\Gamma(\alpha_k)\Gamma(\beta_k)}
\frac{\Gamma(\alpha_k + A_{kn})
\Gamma(\beta_k + B_{kn})
}{\Gamma(\alpha_k + \beta_k + M_{kn})}
\label{eq:cond_Z_one}\\
&\propto
\prod_{k}
\frac{\Gamma(\alpha_k + A_{kn})
\Gamma(\beta_k + B_{kn})
}{\Gamma(\alpha_k + \beta_k + M_{kn})}
\label{eq:cond_Z_two}\\
&=
\prod_{k}
\frac{\Gamma(\alpha_k + A_{kn}^{\neg fn} + z_{fkn}v_{fn})
\Gamma(\beta_k + B_{kn}^{\neg fn} + z_{fkn}\bar{v}_{fn})
}{\Gamma(\alpha_k + \beta_k + M_{kn}^{\neg fn} + z_{fkn})}
\label{eq:cond_Z_three}\\
&\propto
\prod_k
\frac{
\Gamma(\alpha_k + A_{kn}^{\neg fn})
(\alpha_k + A_{kn}^{\neg fn})^{z_{fkn}v_{fn}}
\Gamma(\beta_k + B_{kn}^{\neg fn})
(\beta_k + B_{kn}^{\neg fn})^{z_{fkn}\bar{v}_{fn}}
}
{
\Gamma(\alpha_k + \beta_k + M_{kn}^{\neg fn})
(\alpha_k + \beta_k + M_{kn}^{\neg fn})^{z_{fkn}}
}
\label{eq:app_cond_Z_four}\\
&\propto
\prod_k
\left[
\frac{
(\alpha_k + A_{kn}^{\neg fn})^{v_{fn}}
(\beta_k + B_{kn}^{\neg fn})^{\bar{v}_{fn}}
}
{
(\alpha_k + \beta_k + M_{kn}^{\neg fn})
}
\right]^{z_{fkn}}.
\label{eq:app_cond_Z_five}
\end{align}
The conditional prior term is given by
\begin{align}
p(\bs{z}_{fn} | \bs{Z}_{\neg fn}) = p(\bs{Z})/p(\bs{Z}_{\neg fn}).
\end{align}
Using $p(\bs{Z}) = \prod_{f} \uline{\bs{Z}}_{f}$ and Eq.~\eqref{eq:marginalized_BetaDir_Z} we have
\begin{align}
p(\bs{z}_{fn} | \bs{Z}_{\neg fn}) &\propto
p(\underline{\bs{Z}}_f)
\label{eq:cond_Zprior_two}\\
&
\propto
\prod_k\Gamma(\gamma_k + L_{kn}^{\neg fn} + z_{fkn})
\label{eq:cond_Zprior_four}\\
&=
\prod_k
\Gamma(\gamma_k + L_{kn}^{\neg fn})
(\gamma_k + L_{kn}^{\neg fn})^{z_{fkn}}
\label{eq:cond_Zprior_five}\\
&\propto
\prod_k
(\gamma_k + L_{kn}^{\neg fn})^{z_{fkn}}.
\label{eq:cond_Zprior_six}
\end{align}
Using $\sum_{k} p(\bs{z}_{fn} = \ve{e}_{k} | \bs{Z}_{\neg fn}) =1$, a simple closed-form expression of $p(\bs{z}_{fn} | \bs{Z}_{\neg fn})$ is obtained as follows:
\begin{align}
p(\bs{z}_{fn} = \ve{e}_{k}  | \bs{Z}_{\neg fn}) &= \frac{\gamma_k + L_{kn}^{\neg fn}}{\sum_{k} (\gamma_k + L_{kn}^{\neg fn})} \\
& = \frac{\gamma_k + L_{kn}^{\neg fn}}{\sum_{k} \gamma_k + N-1}.
\end{align}

Combining Eqs.~\eqref{eqn:cond1}, \eqref{eq:app_cond_Z_five} and \eqref{eq:cond_Zprior_six}, we obtain
\begin{align}
p(\bs{z}_{fn} | \bs{Z}_{\neg fn},\bs{V}) \propto
\prod_k
\left[
(\gamma_k + L_{fk}^{\neg fn} ) 
\frac{(\alpha_k + A_{kn}^{\neg fn})^{v_{fn}}
(\beta_k + B_{kn}^{\neg fn})^{\bar{v}_{fn}}
}{\alpha_k + \beta_k + M_{kn}^{\neg fn}}
\right]^{z_{fkn}}.
\end{align}

\section{\new{Alternative Gibbs sampler for the Dir-Dir model}}\label{sec:appendix_DirDir}

\new{
In this appendix, we show how to derive an alternative Gibbs sampler based on a single augmentation, like in the \texttt{Beta-Dir} model. This is a conceptually interesting result, though it does not lead to an efficient implementation.}

\new{
Likewise the \texttt{Beta-Dir} model, the \texttt{Dir-Dir} model can be augmented using the single indicator variables $\bs{z}_{fn}$, as follows:
\begin{align}
     \bs{h}_{n} &\sim \text{Dirichlet}(\bs{\eta})\\
    \bs{w}_{f} & \sim \text{Dirichlet}(\bs{\gamma}) \\
    \bs{z}_{fn} | \bs{w}_f & \sim \text{Discrete}(\bs{w}_f) \\
    v_{fn} | \bs{h}_{n}, \bs{z}_{fn} & \sim \text{Bernoulli}\left(\prod_k h_{kn}^{z_{fkn}}\right) 
\end{align}
Note that compared to Eqs.~\eqref{eq:betadir-aug1}-\eqref{eq:betadir-aug4} only the prior on $\bs{h}_{n}$ is changed.}

\new{
Like in \texttt{Beta-Dir}, we seek in this appendix to derive a Gibbs sampler from the conditional probabilities $p(\bs{z}_{fn} | \bs{Z}_{\neg fn}, \bs{V})$ given by
\begin{align}
p(\bs{z}_{fn} | \bs{Z}_{\neg fn}, \bs{V}) \propto p(\bs{V} | \bs{Z})p(\bs{z}_{fn} | \bs{Z}_{\neg fn}).
\end{align} 
The conditional prior term is identical to that of \texttt{Beta-Dir} and given by
\begin{align}
p(\bs{z}_{fn} | \bs{Z}_{\neg fn}) \propto
\prod_k (\gamma_k + L_{kn}^{\neg fn})^{z_{fkn}}.
\end{align}
Like in \texttt{Beta-Dir}, the likelihood term factorizes as $p(\bs{V} | \bs{Z}) = \prod_{n} p(\bs{v}_n | \bs{Z}_n)$, and we now derive the expression of $p(\bs{v}_n | \bs{Z}_n)$. As compared to \texttt{Beta-Dir}, a major source of difficulty lies in the fact that $p(\ve{h}_n)$ does not fully factorize anymore because of the Dirichlet assumption (and in particular $\sum_k h_{kn}=1$). In the following, we use the multinomial theorem to obtain Eq.~\eqref{eq:multinomth}\footnote{Many thanks to Xi'an (Christian Robert) for giving us the trick via StackExchange.} and we use the expression of the normalization constant of the Dirichlet distribution to obtain Eq.~\eqref{eq:app_pV_Z}:
\begin{align}
p(\bs{v}_n | \bs{Z}_n) &= 
\int
p(\bs{h}_n)
\prod_{f}
p(v_{fn} | \bs{h}_n, \bs{z}_{fn})
 \,d\mathbf{h}_n\\
&=\int
\frac{\Gamma(\sum_k \eta_k)}{\prod_k \Gamma(\eta_k)}
\prod_k h_{kn}^{\eta_k-1}
\prod_{f}\prod_k
\left[h_{kn}^{v_{fn}} (1-h_{kn})^{1-v_{fn}}\right]^{z_{fkn}}
 \,d\mathbf{h}_n
\\&
= \frac{\Gamma(\sum_k \eta_k)}{\prod_k \Gamma(\eta_k)}
\int
\prod_k h_{kn}^{\eta_n + A_{kn}-1}
(1-h_{kn})^{B_{kn}}
 \,d\mathbf{h}_n \\
&=
\frac{\Gamma(\sum_k \eta_k)}{\prod_k \Gamma(\eta_k)}
\int
\prod_k h_{kn}^{\eta_n + A_{kn}-1}
\sum_{j_k=0}^{B_{kn}}
\binom{B_{kn}}{j_k}
(-h_{kn})^{j_k}
 \,d\mathbf{h}_n
 \label{eq:multinomth}
 \\&
=\frac{\Gamma(\sum_k \eta_k)}{\prod_k \Gamma(\eta_k)}
\int
\sum_{j_1=0}^{B_{1n}}
...
\sum_{j_K=0}^{B_{Kn}}
\prod_k h_{kn}^{\eta_k + A_{kn}-1}
\binom{B_{kn}}{j_k}
(-h_{kn})^{j_k}
 \,d\mathbf{h}_n
 \\&
= \frac{\Gamma(\sum_k \eta_k)}{\prod_k \Gamma(\eta_k)}
\sum_{j_1=0}^{B_{1n}}
...
\sum_{j_K=0}^{B_{Kn}}
\prod_k
(-1)^{j_k}\binom{B_{kn}}{j_k}
\int
\prod_k h_{kn}^{\eta_k + A_{kn} + j_k -1}
 \,d\mathbf{h}_n
\\&
= \frac{\Gamma(\sum_k \eta_k)}{\prod_k \Gamma(\eta_k)}
\sum_{j_1=0}^{B_{1n}}
...
\sum_{j_K=0}^{B_{Kn}}
\prod_k
(-1)^{j_k} \binom{B_{kn}}{j_k}
\frac{\Gamma(\eta_k + A_{kn} + j_k)}
{\Gamma(\sum_k \eta_k + A_{kn} + j_k)}. 
\label{eq:app_pV_Z}
 \end{align}
We conclude that, though available in closed form, the expression of $p(\bs{v}_n | \bs{Z}_n)$ (and thus $p(\bs{z}_{fn} | \bs{Z}_{\neg fn})$) involves the computation of $K\prod_{k=1}^K B_{kn}$ terms involving binomial coefficients, which is impractical in typical problem dimensions.
}

\end{document}